\documentclass{article} 
\usepackage{iclr2025_conference,times}

\usepackage{amsmath,amsfonts,bm}

\def\eqref#1{equation~\ref{#1}}

\def\1{\bm{1}}

\DeclareMathAlphabet{\mathsfit}{\encodingdefault}{\sfdefault}{m}{sl}
\SetMathAlphabet{\mathsfit}{bold}{\encodingdefault}{\sfdefault}{bx}{n}

\usepackage{hyperref}
\usepackage{url}

\usepackage[utf8]{inputenc} %
\usepackage[T1]{fontenc}    %
\usepackage{amsfonts}       %
\usepackage{nicefrac}       %
\usepackage{microtype}      %
\usepackage{arydshln} 
\usepackage{tabularray}
\usepackage{graphicx}
\usepackage{amsmath,amssymb}
\usepackage{amsthm}
\usepackage{wrapfig}
\usepackage{multirow}
\usepackage{multicol}
\usepackage{color,xcolor}
\usepackage[caption=false]{subfig}
\usepackage{booktabs}
\usepackage{threeparttable}
\usepackage{courier}
\usepackage{adjustbox}
\usepackage{supertabular}
\usepackage{stmaryrd}
\usepackage{makecell}
\usepackage{marvosym}
\usepackage{colortbl}
\usepackage{pifont}
\usepackage{appendix}
\colorlet{darkgreen}{green!65!black}
\colorlet{darkblue}{blue!75!black}
\colorlet{darkred}{red!80!black}
\definecolor{lightblue}{HTML}{0071bc}
\definecolor{lightgreen}{HTML}{39b54a}
\usepackage{caption}
\usepackage{times}
\usepackage{epsfig}
\usepackage{graphicx}
\usepackage{verbatim}
\usepackage{bbm}

\usepackage{enumitem,kantlipsum}

\newcolumntype{g}{>{\columncolor{tbgray}}c}

\definecolor{red}{rgb}{0.8, 0.0, 0.0}
\definecolor{green}{rgb}{0.0, 0.5, 0.0}
\definecolor{tbgray}{gray}{.92}

\title{Revisit Large-Scale Image-Caption Data in Pre-training Multimodal Foundation Models}

\author{Zhengfeng Lai\thanks{Equal contribution.}, \,Vasileios Saveris\footnotemark[1], \,Chen Chen, \,Hong-You Chen, \,Haotian Zhang \\
\textbf{Bowen Zhang, \,Juan Lao Tebar, \,Wenze Hu, \,Zhe Gan, \,Peter Grasch} \\ 
\textbf{Meng Cao, \,Yinfei Yang}\thanks{Corresponding author.} \\
Apple AI/ML \\
\small{\texttt{\{jeff\_lai,v\_saveris,pgrasch,mengcao,yinfeiy\}@apple.com}}
}

\iclrfinalcopy 
\begin{document}

\maketitle

\begin{abstract}
Recent advancements in multimodal models highlight the value of rewritten captions for improving performance, yet key challenges remain. For example, while synthetic captions often provide superior quality and image-text alignment, it is not clear whether they can fully replace AltTexts: the role of synthetic captions and their interaction with original web-crawled AltTexts in pre-training is still not well understood. Moreover, different multimodal foundation models may have unique preferences for specific caption formats, but efforts to identify the optimal captions for each model remain limited. In this work, we propose a novel, controllable, and scalable captioning pipeline designed to generate diverse caption formats tailored to various multimodal models. By examining Short Synthetic Captions (SSC) towards Dense Synthetic Captions (DSC+) as case studies, we systematically explore their effects and interactions with AltTexts across models such as CLIP, multimodal LLMs, and diffusion models. Our findings reveal that a hybrid approach that keeps both synthetic captions and AltTexts can outperform the use of synthetic captions alone, improving both alignment and performance, with each model demonstrating preferences for particular caption formats. This comprehensive analysis provides valuable insights into optimizing captioning strategies, thereby advancing the pre-training of multimodal foundation models.
\end{abstract}

\section{Introduction}
\begin{figure}[ht]
    \centering
    \includegraphics[width=1\linewidth]{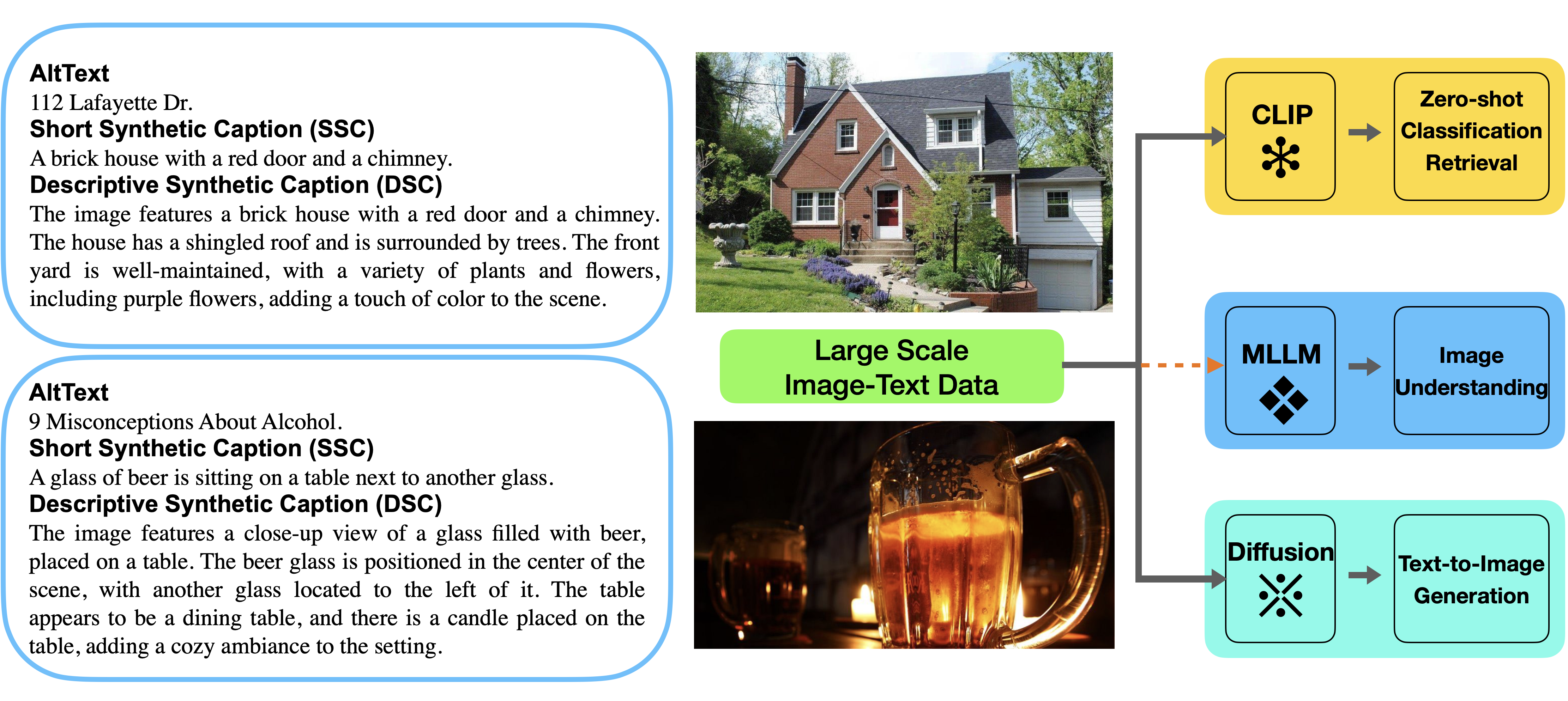}
    \caption{The role of image-text data in multimodal foundation models: a \textbf{key} component in training CLIP and Diffusion Model, and \textbf{essential} for multimodal LLM (MLLM) pre-training alongside text and interleaved image-text data. We propose a controllable captioning pipeline to synthesize different types of captions and explore optimal image-text data recipes for training these foundation models.  }
    \label{fig:example_caption}
\end{figure}

Large-scale image-text datasets have been crucial in advancing multimodal foundation models. For instance, CLIP~\citep{CLIP} is pre-trained on 400 million image-text pairs collected from the Web. However, web-crawled data, particularly AltText, often suffer from insufficient visual details and noisy content, as illustrated in Fig.~\ref{fig:example_caption}. Recent studies highlight the benefits of synthetic captions, which provide better image-text alignment and improved data quality. Research on LaCLIP~\citep{laclip} and ShareGPT4V~\citep{sharegpt4v} demonstrates that synthetic captions can improve the performance of CLIP and multimodal large language models (MLLMs), respectively. This raises a key question: if higher-quality synthetic captions can be generated, could they fully replace web-crawled AltText? Should we consider disregarding AltText altogether?

\begin{figure}[t]
    \centering
    \begin{minipage}[b]{0.5\textwidth}
        \includegraphics[width=\linewidth]{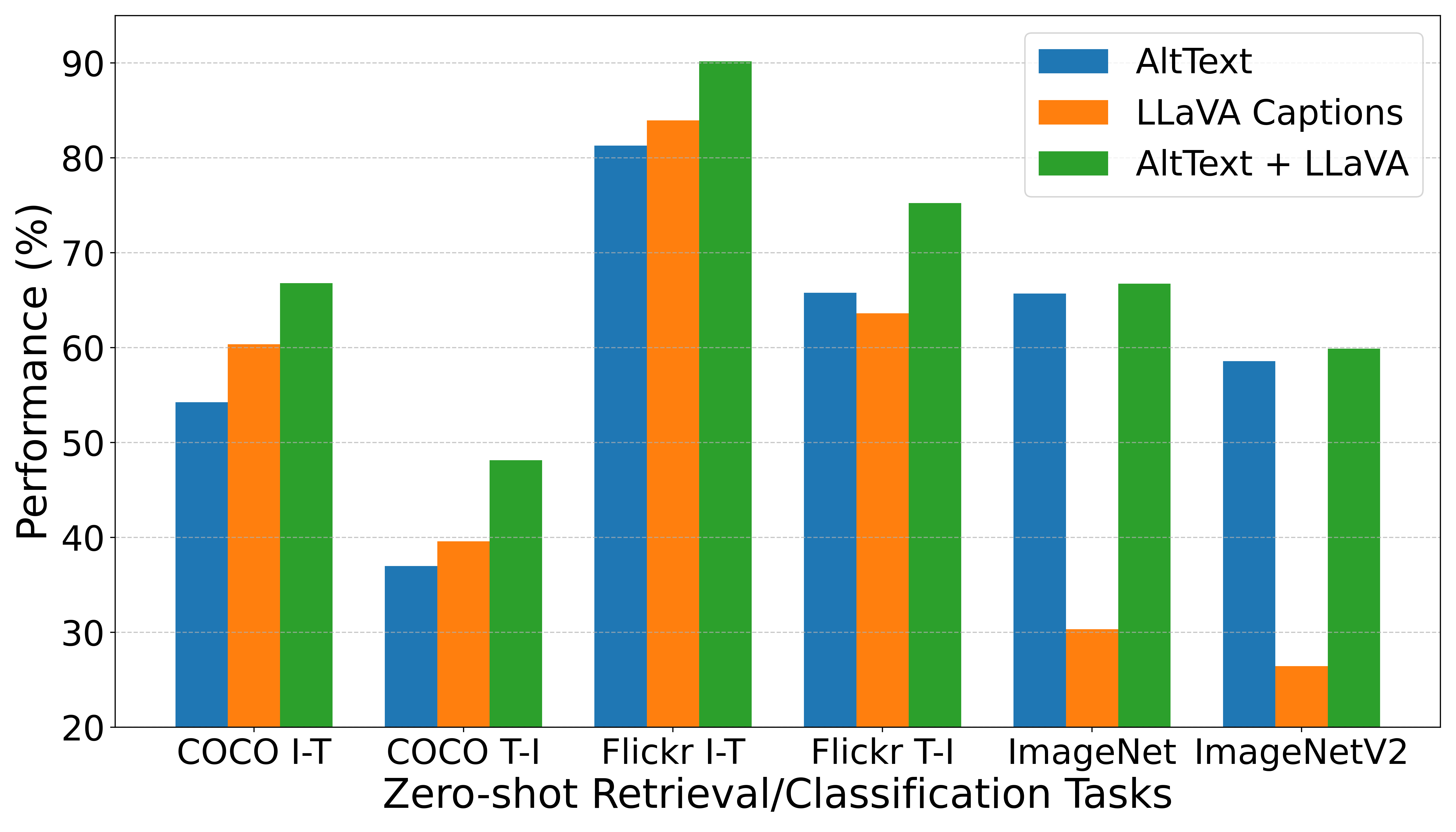}
        \centerline{{(a)}}\medskip
    \end{minipage}\hfill
    \begin{minipage}[b]{0.5\textwidth}
        \includegraphics[width=\linewidth]{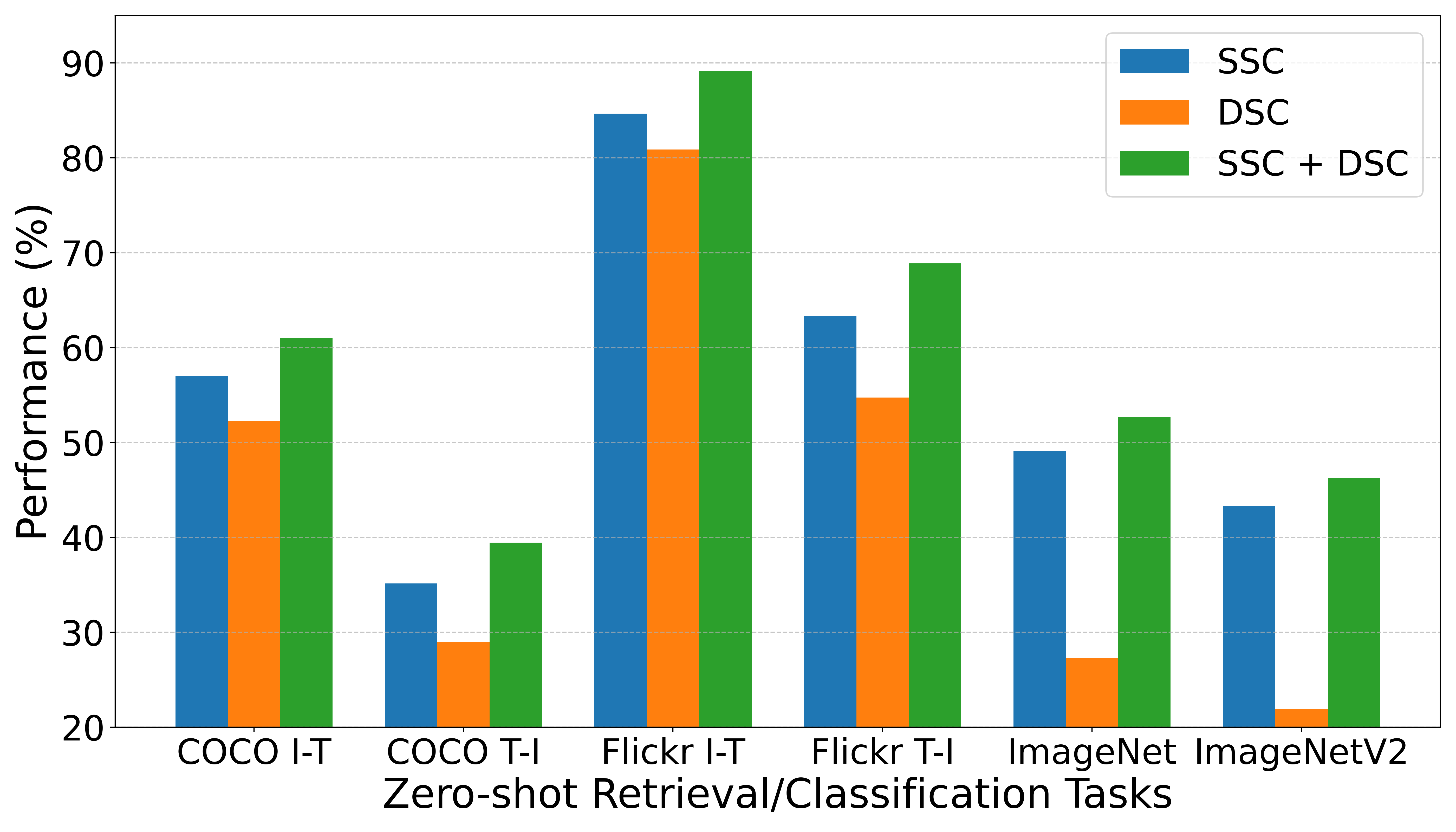}
        \centerline{{(b)}}\medskip
    \end{minipage}
    \caption{Zero-shot retrieval and classification performance of CLIP models. (a) The effect of synthetic captions (LLaVA recaptioned) and AltText: solely using LLaVA captions can improve retrieval tasks but significantly deteriorate the zero-shot classification performance. (b) The effect of different formats of synthetic captions on CLIP: Short Synthetic Captions (SSC) show superior results to Descriptive Synthetic Captions (DSC) and the combination of them achieves the best results. 
    }
    \label{fig:fig1}
\vspace{-5mm}
\end{figure}

To investigate this question, we first adopt the approach of VeCLIP~\citep{veclip} and train CLIP using synthetic captions generated by LLaVA~\citep{llava}. Similar to the results discussed in~\citet{li2024if}, training CLIP fully on synthetic captions of higher quality degrades CLIP's performance significantly: as shown in Fig.~\ref{fig:fig1}, when compared to using only AltText, the use of LLaVA captions results in a substantial drop on zero-shot ImageNet classification tasks. However, after combining original noisy AltText and LLaVA captions, we achieve the best results in both classification and retrieval tasks. This observation raises a critical question: what constitutes the optimal image-text data for multimodal foundation models? Despite its importance, research on the interplay between synthetic captions and AltText remains limited. Our findings suggest that while rewriting techniques can enhance image-text alignment, they may reduce data diversity due to dependence on a limited set of LLMs or MLLMs for caption generation. Specifically, since CLIP is a foundational vision model that benefits from learning diverse concepts, relying on synthetic captions can potentially hinder CLIP's training due to lack of diversity in vocabulary and mentioned concepts~\citep{laclip}.

In addition to the role of AltText, another open question concerns the optimal formats for synthetic captions. For instance, advanced multimodal models like LLaVA-NeXT~\citep{llavanext-ablations} indicate that recaptioned datasets are advantageous during stages focused on high-quality knowledge acquisition. DALL-E 3~\citep{dalle3} demonstrates that using 95\% synthetic captions can yield superior results, particularly when the captions are highly descriptive. Similarly, MM1~\citep{mm1} shows that even a small fraction (7\%) of high-quality caption data can significantly boost few-shot performance. Given these insights, our work focuses on two key unresolved questions: \textit{1) What is the role and value of synthetic captions, and how do they interact with the original AltText? 2) What types of synthetic captions are most effective for different foundation models?} To address the first question, we revisit why prior works~\citep{dalle3,veclip,mm1} continue using noisy web-crawled AltText, even when rewritten captions are available during training. Intuitively, since CLIP is a straightforward model pre-trained on image-text pairs, highly aligned captions should be advantageous.  However, relying solely on synthetic captions may actually degrade CLIP’s performance, as shown in Fig.~\ref{fig:fig1}(a).
For the second question, we investigate the effects of Short Synthetic Captions (SSC) and Descriptive Synthetic Captions (DSC) on CLIP. As depicted in Fig.~\ref{fig:fig1}(b), surprisingly, more descriptive captions yield inferior results compared to shorter captions for CLIP training, despite their greater detail.

To explore these insights further to address the two questions, we introduce a novel, controllable, and scalable captioning pipeline that enables the generation of diverse caption formats at scale, tailored to the specific needs of different multimodal foundation models. Our pipeline is designed to build large-scale image-text data for the pre-training stage in a scalable way. 
With this pipeline, we use SSC and DSC as two main examples on how to customize the captioning format. Our pipeline can serve as a cost-effective alternative to GPT-4V for generating high-quality captions. To solve the second question, we conduct a comprehensive study on the effectiveness of different types of synthetic captions across a range of foundational models and downstream tasks. Our approach involves a systematic evaluation of various captioning strategies, including SSC, DSC, and mixed training methods that combine original AltText with synthetic data. We seek to determine the optimal captioning techniques for specific models, such as CLIP, diffusion models, and multimodal LLMs, and to assess their impact on both model performance and data diversity. Furthermore, we investigate the interaction between synthetic captions and original AltText, analyzing whether a hybrid approach can balance the need for diverse data with the benefits of enhanced image-text alignment. 

Overall, our contributions are summarized as follows.
\begin{itemize}[topsep=0pt,leftmargin=*]
    \item We explore the MLLM as the image describer and present a controllable and human-aligned captioning pipeline to convert MLLM into an image captioner. 
    \item We synthesize several formats of captions including Short Synthetic Captions (SSC) towards Dense Synthetic Captions (DSC+), then conduct extensive pre-training experiments to systematically study the role of synthetic captions and their intersection with original AltText across three multimodal foundation models. 
    \item We verify the image-caption training recipe that 1) AltText provide data variety and synthetic captions provide better image-text alignment, 2) different foundation models have their own preferred formats, which highlights the necessity and importance of the controllable captioning pipeline in building multimodal foundation models. 
\end{itemize}

\section{Related Work}
\textbf{Multimodal Foundation Models.} CLIP~\citep{CLIP} is one of the pioneering multimodal foundation models connecting images and text. By training on 400 million image-text pairs, CLIP shows strong zero-shot image classification and retrieval capabilities. It lays the groundwork for the development of more advanced multimodal foundation models, such as multimodal large language models (MLLMs)~\citep{llava, cogvlm, internvl, cambrian, zhang2024mm15methodsanalysis} for vision-language understanding and diffusion models~\citep{SD, SDXL} for text-to-image generation.  These advanced models often utilize CLIP's vision tower as their vision encoder. 

\textbf{Improving Image-Text Data.} Web-crawled image-text data often suffer from issues like image-text misalignment and poor-quality textual descriptions~\citep{veclip,li2024if}. There are two common ways for improving image-text data: 1) data filtering based methods remove low-quality data such as misaligned image-text pairs by human-assisted systems~\citep{yu2024rlhf,sun2023aligning} or pre-trained models~\citep{li2022blip,laion,datacomp,DFN}; 2) data recaptioning based methods usually leverage a LLM to rewrite the original caption or a MLLM to rewrite a caption for the image. For example,  ShareGPT4V~\citep{sharegpt4v} uses GPT-4V to write highly descriptive captions for their images. LaCLIP~\citep{laclip} leverages several LLMs to rewrite captions with different writing styles for data diversity. Recap-DataComp-1B~\citep{li2024if} uses a LLaMA-3 based model to scale the captions. Different from the aforementioned works, we mainly focus on generating different types of captions and exploring 1) the format of ideal captions needed for each multimodal foundation model and 2) a systematic analysis of the intersection between AltText and synthetic captions. 
\section{Customized Re-captioning for Multimodal Foundation Models}

\begin{figure}[t]
    \centering
    \includegraphics[width=0.98\linewidth]{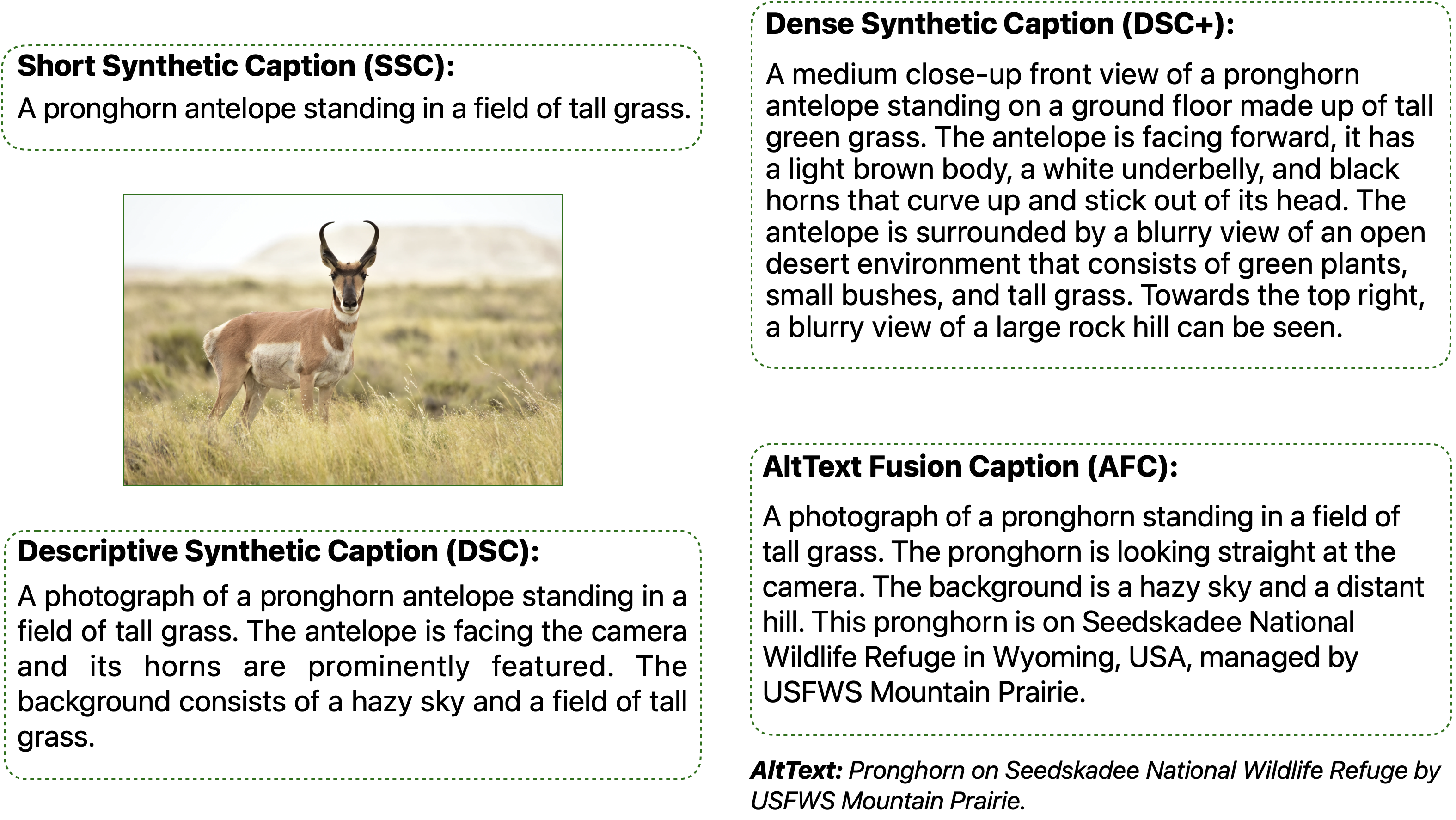}
    \includegraphics[width=0.98\linewidth]{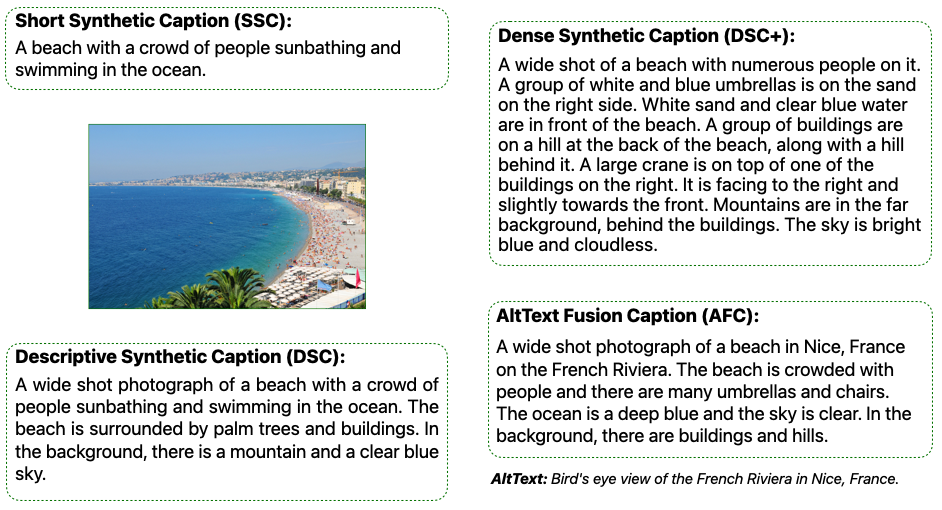}
    \vspace{-0.2cm}
    \caption{Examples of controllable captions of diverse formats generated by our captioner: we can generate from brief to dense descriptions and fuse AltText into the caption (AFC). }
    \label{fig:captioner_examples}
\end{figure}

Image-text data are fundamental to the success of multimodal foundation models, serving as a bridge between visual and textual representations. 
For example, CLIP~\citep{CLIP} is pre-trained on 400M web-crawled image-text pairs, enabling it to learn rich, transferable representations that can be applied to various downstream tasks. 
The importance of precise and detailed captioning is further highlighted in LLaVA-NeXT~\citep{llavanext-ablations}, where re-captioned detailed descriptions are utilized for model training at Stage-1.5, enhancing the model's ability to understand and generate nuanced content. Similarly, DALL-E 3~\citep{dalle3} shows that the prompt-following capabilities of text-to-image models can be significantly improved by training on highly-descriptive generated image captions. This shows the critical role of captions in shaping a model's capacity to align visual and textual information, ultimately improving performance across a wide range of multimodal tasks. However, the optimal captioning strategy for different foundational models remains under-explored. To address this gap, we introduce a novel, controllable, and scalable captioning pipeline designed to generate diverse caption formats at scale, supported by evaluation metrics that ensure high CLIP scores and minimal hallucination. 
We summarize the capability of our captioning model by generating the following formats of captions as shown in Fig.~\ref{fig:captioner_examples}: 

\begin{itemize}[topsep=0pt,leftmargin=*]
    \item \textbf{Short Synthetic Caption (SSC)}: a concise sentence that describes the primary subject of the image.
    \item \textbf{Descriptive Synthetic Caption (DSC)}: a description limited to 78 tokens, emphasizing the central subject and key visual elements.
    \item \textbf{Dense Synthetic Caption (DSC+)}:  a more comprehensive description detailing the main subject along with the background, setting, and any significant objects or actions.
    \item \textbf{AltText Fusion Caption (AFC)}: a caption similar to DSC, but integrated with AltText where appropriate. This type of caption removes unnecessary details often found in AltText, offering a cleaner, more cohesive description than a simple concatenation of AltText and synthetic caption.
\end{itemize}

\subsection{MLLM as An Image Describer}

\begin{figure}[t]
    \centering
    \includegraphics[width=0.98\linewidth]{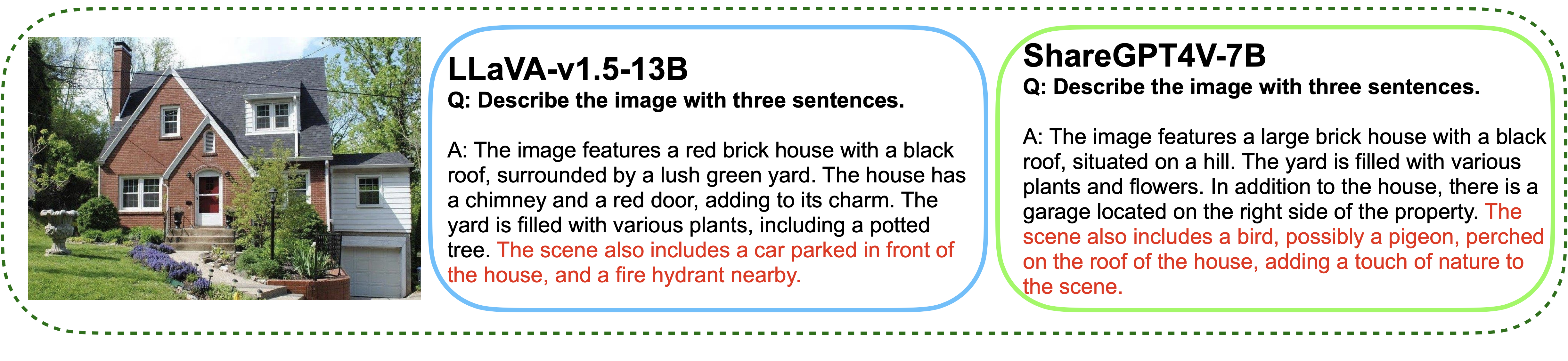}
    \vspace{-0.3cm}
    \caption{Directly using MLLMs as image captioners may result in hallucinations and generate captions that do not align with specific instructions: both LLaVA~\citep{llava} and ShareGPT4V~\citep{sharegpt4v} generate over three sentences and obvious hallucination.  }
    \label{fig:exist_captioner}
\end{figure}

VeCLIP~\citep{veclip} employs LLaVA~\citep{llava} for image captioning, while ShareGPT4V~\citep{sharegpt4v} utilizes GPT-4V for this task. 
Compared to traditional image captioners like BLIP~\citep{blip}, MLLMs offer several advantages and are natually good image describers. MLLMs can generate longer and more detailed captions, as demonstrated by LLaVA-NeXT~\citep{llavanext-ablations}, where a 34B model was used to produce highly descriptive captions. This ability to generate more context-aware descriptions stems from the integration of a large language model (LLM) with a vision encoder. This combination allows MLLMs to capture fine-grained visual details and complex inter-object relationships. Additionally, MLLMs benefit from their multi-stage training process, which combines pre-training on large-scale datasets and supervised fine-tuning for downstream tasks. These characteristics make MLLMs a powerful tool for generating more descriptive captions. 
However, directly using MLLM as an image describer may have two major limitations: 1) MLLM may not strictly follow the instruction to generate a specific format of caption~\citep{liu2023mitigating}; 2) these instruction fine-tuned MLLMs tend to hallucinate. As shown in Fig.~\ref{fig:exist_captioner}, both LLaVA~\citep{llava} and ShareGPT4V~\citep{sharegpt4v} fail to describe the image using only three sentences but generate hallucinated contents (highlighted in red). 
Although GPT-4V shows stronger capability and many works use it for captioning, the scalability remains limited due to the cost. Therefore, in this work, we focus on building a cost-effective captioner instead of using GPT-4V. 

To alleviate the above two limitations, we first investigate the origins of hallucination in MLLMs, hypothesizing two primary sources: 1) inherent limitations of the LLM, and 2) the quality of supervised fine-tuning (SFT) datasets, which are often themselves synthetically derived or processed. We focus on the latter, proposing that mitigating hallucinations at the dataset level is essential for converting an MLLM into an effective captioner. We also address the format-following issue by fine-tuning the MLLM on a curated captioning-specific dataset, transforming it into a purpose-built captioning model to generate diverse captions.

\subsection{Two-Stage Human-Aligned Captioning}

\textbf{Stage 1: transforming MLLM into a customized captioner.} To minimize hallucinations from MLLMs, we begin by constructing a clean and precise fine-tuning dataset. Instead of relying on GPT-4 generated data, we curate a high-quality dataset of human-annotated image-text pairs, named Stage-1-1M. This dataset contains short, human-curated captions, with five captions per image. Additionally, we integrate an OCR detection model to extract in-image text, which, alongside the captions, is fed into an LLM for summarization and controlled rewriting. By employing a strict prompt, we prevent the LLM from introducing extraneous information, while few-shot prompts guide the model in generating various caption formats, including both concise and descriptive styles. We further enhance the dataset through post-processing, using heuristic and model-based quality checks to improve its overall quality. This comprehensive fine-tuning on the Stage-1-1M dataset transforms the MLLM, specifically the 3B version of MM1~\citep{mm1}, into a customized captioner that aligns closely with the intended output characteristics.

\begin{figure}[t]
    \centering
    \includegraphics[width=0.94\linewidth]{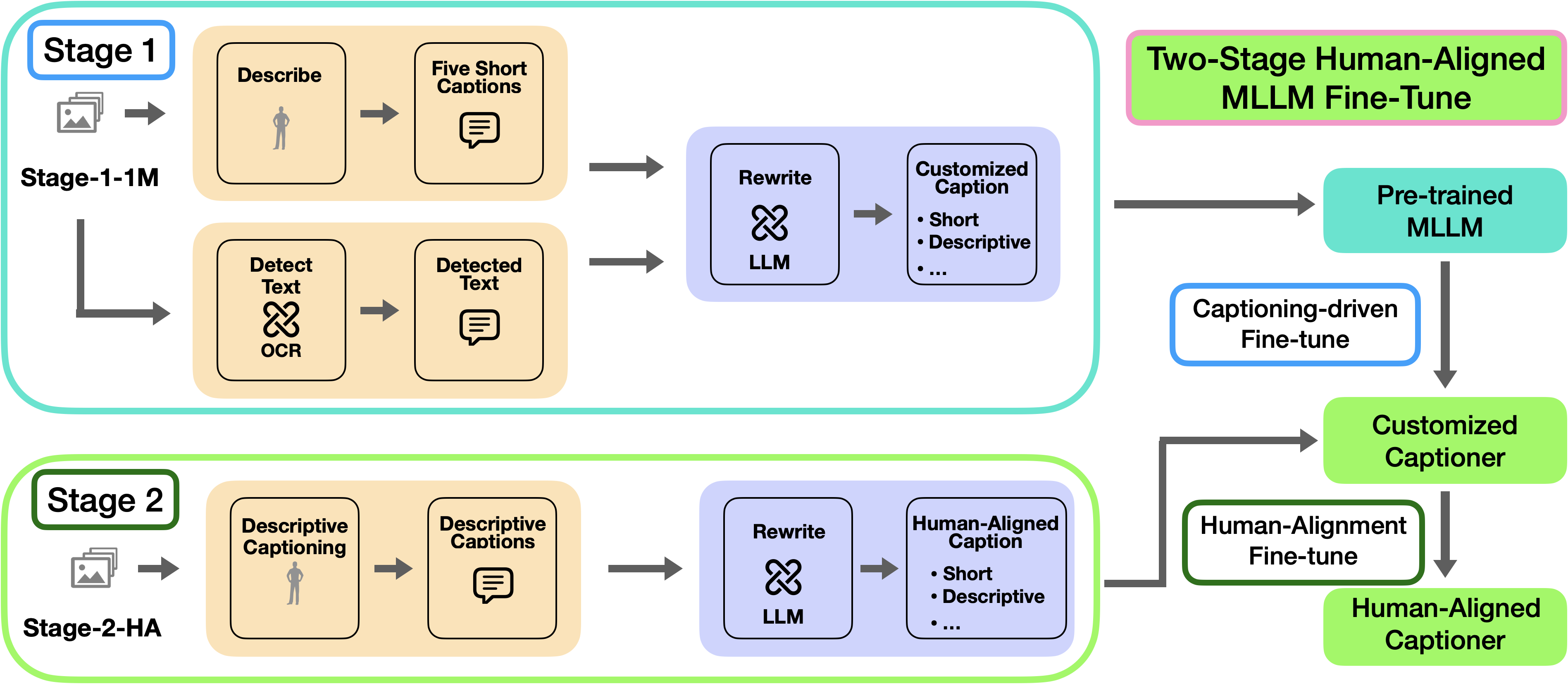}
    \vspace{-0.1cm}
    \caption{Overview of the controllable and human-aligned captioning pipeline. In Stage 1, we convert a pre-trained MLLM into a customized captioner that strictly follows the captioning instructions. In Stage 2, we leverage human-aligned captions to further fine-tune the captioner. }
    \label{fig:captioner}
\end{figure}

\textbf{Stage 2: Human-aligned further fine-tuning.} 
While the Stage-1-1M dataset effectively establishes a foundation, it lacks the depth required for more descriptive captioning tasks. In Stage 2, we address this by incorporating descriptive human-annotated data to enhance caption diversity and quality. We curate a new dataset, named Stage-2-HA, specifically designed for detailed captioning. This dataset is meticulously annotated by human experts to capture nuanced visual elements and complex scene descriptions. Post-annotation, we leverage an LLM to reformat these captions into multiple stylistic variations, including both concise and richly descriptive formats. By imposing strict constraints during the LLM processing, we ensure alignment with human-generated content, avoiding the pitfalls of hallucination. The captioner is then fine-tuned on the Stage-2-HA dataset, resulting in a highly refined and human-aligned captioning model capable of generating captions tailored to specific use cases. This dual-stage fine-tuning process not only enhances the model’s adaptability but also ensures a balance between brevity and descriptiveness, making it a versatile tool in controllable captioning. The overview of the complete captioning pipeline is shown in Fig.~\ref{fig:captioner}.

\subsection{Caption Analysis}

\begin{figure*}[t!]
\begin{center}
	\begin{tabular}{cc}
	\includegraphics[width=0.4\textwidth]{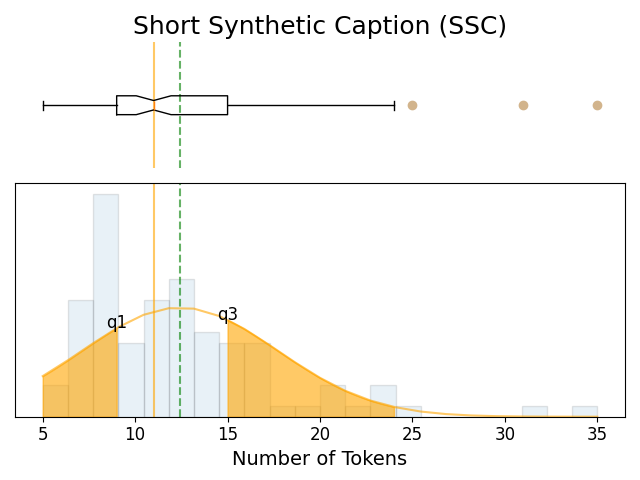}  &
	\includegraphics[width=0.4\textwidth]{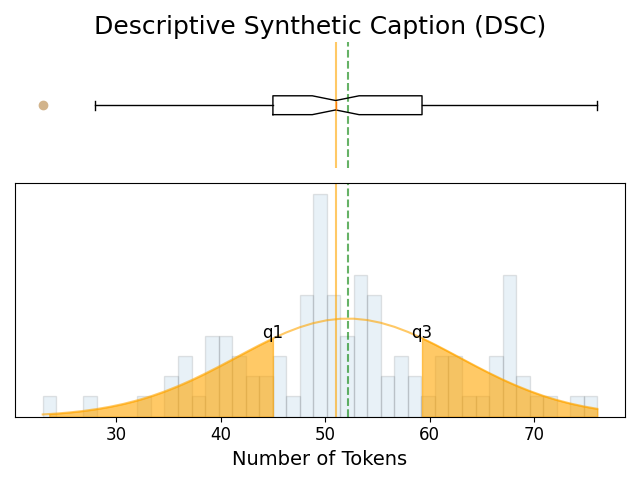} 
 \\
 
	\includegraphics[width=0.4\textwidth]{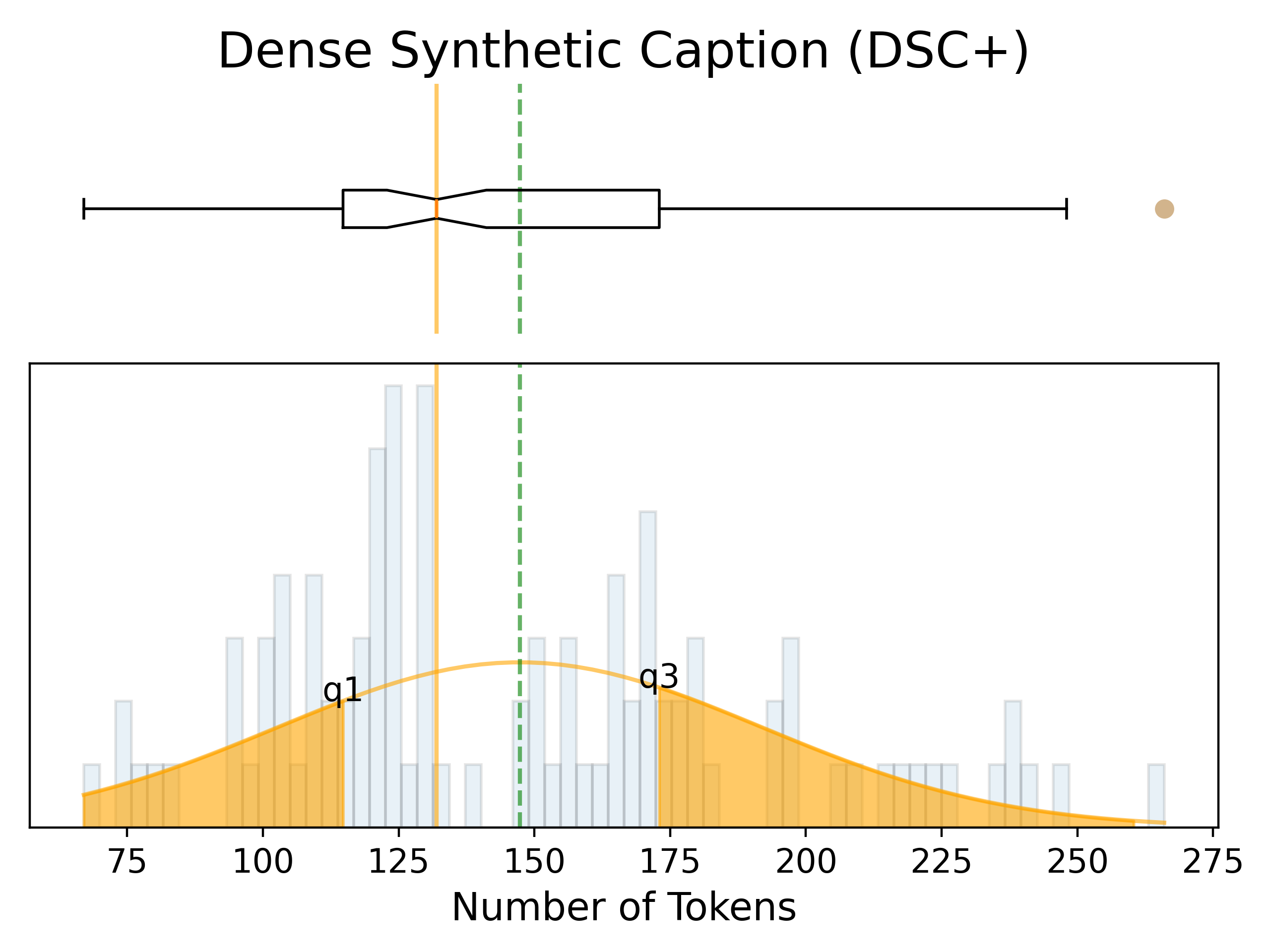}  &
	\includegraphics[width=0.4\textwidth]{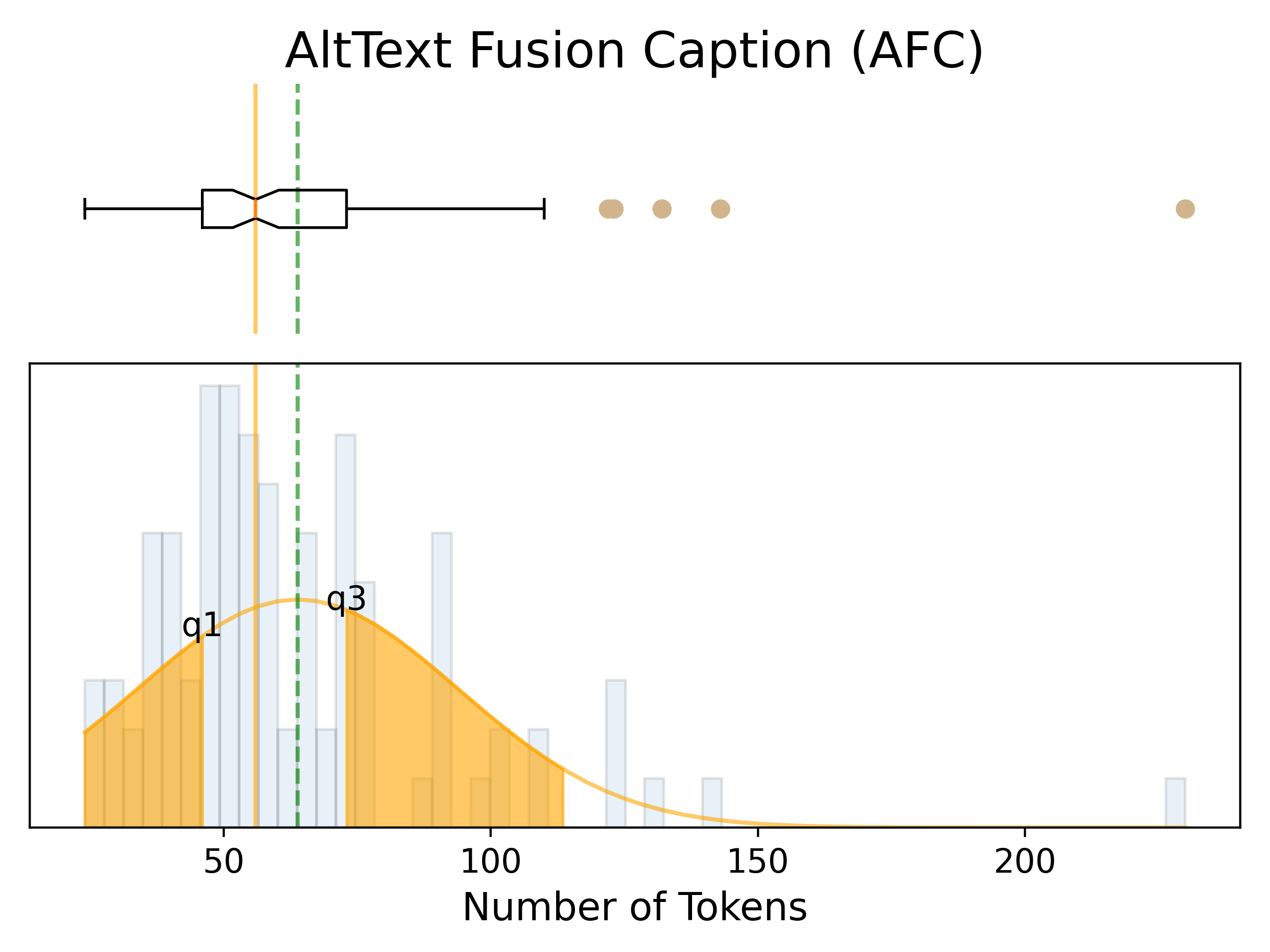}  \\
	\end{tabular}
\end{center}
\vspace{-0.5cm}
\caption{Distribution of token lengths of our generated captions in four formats: we caption COCO 2017 images and visualize their distributions.}
\label{fig:captions_lengths}
\end{figure*}

\begin{figure*}[t]
\begin{center}
	\begin{tabular}{cccc}
	\hspace{-6mm}  \includegraphics[width=0.27\textwidth]{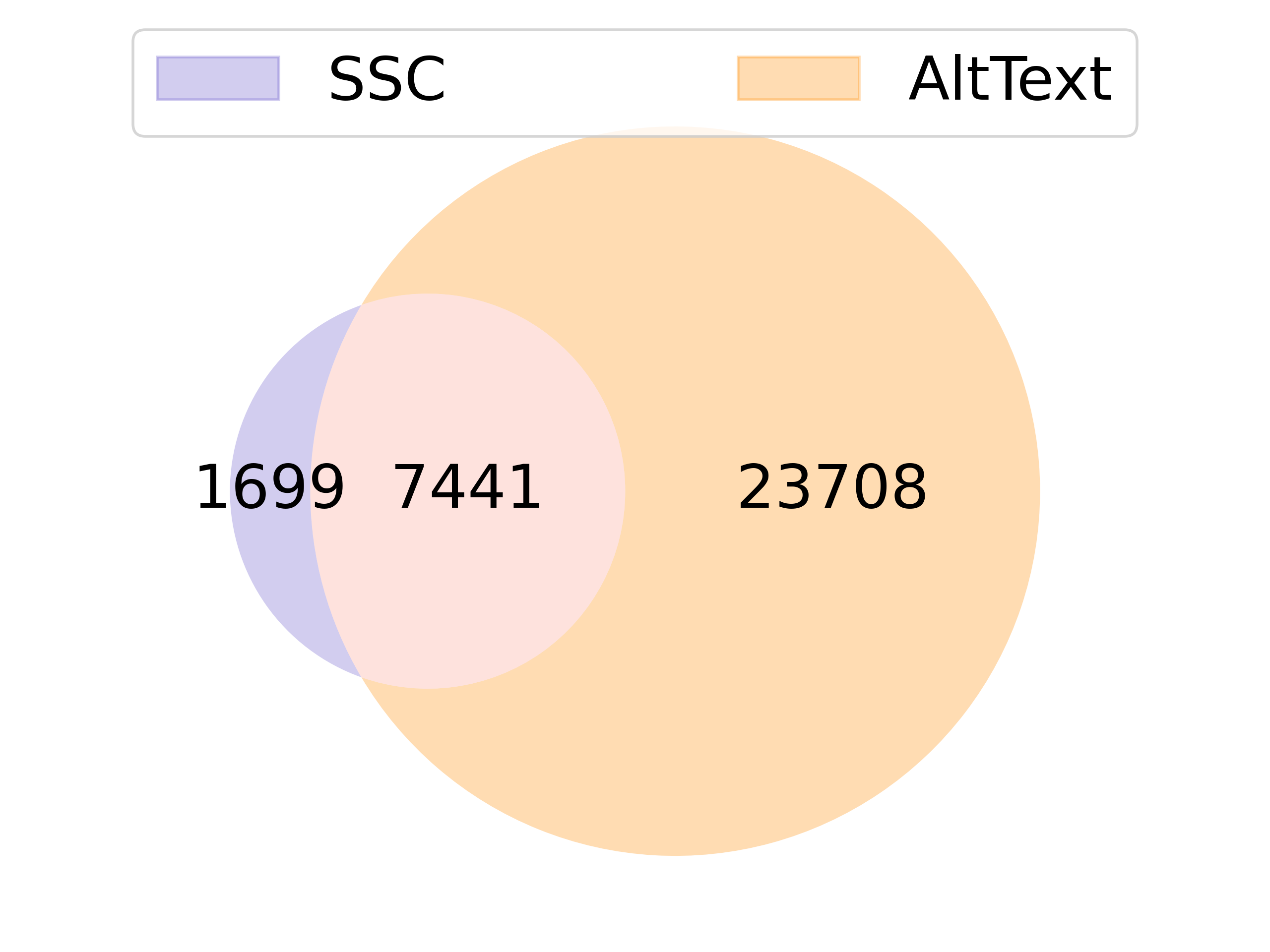} \hspace{-8mm} &
	\includegraphics[width=0.27\textwidth]{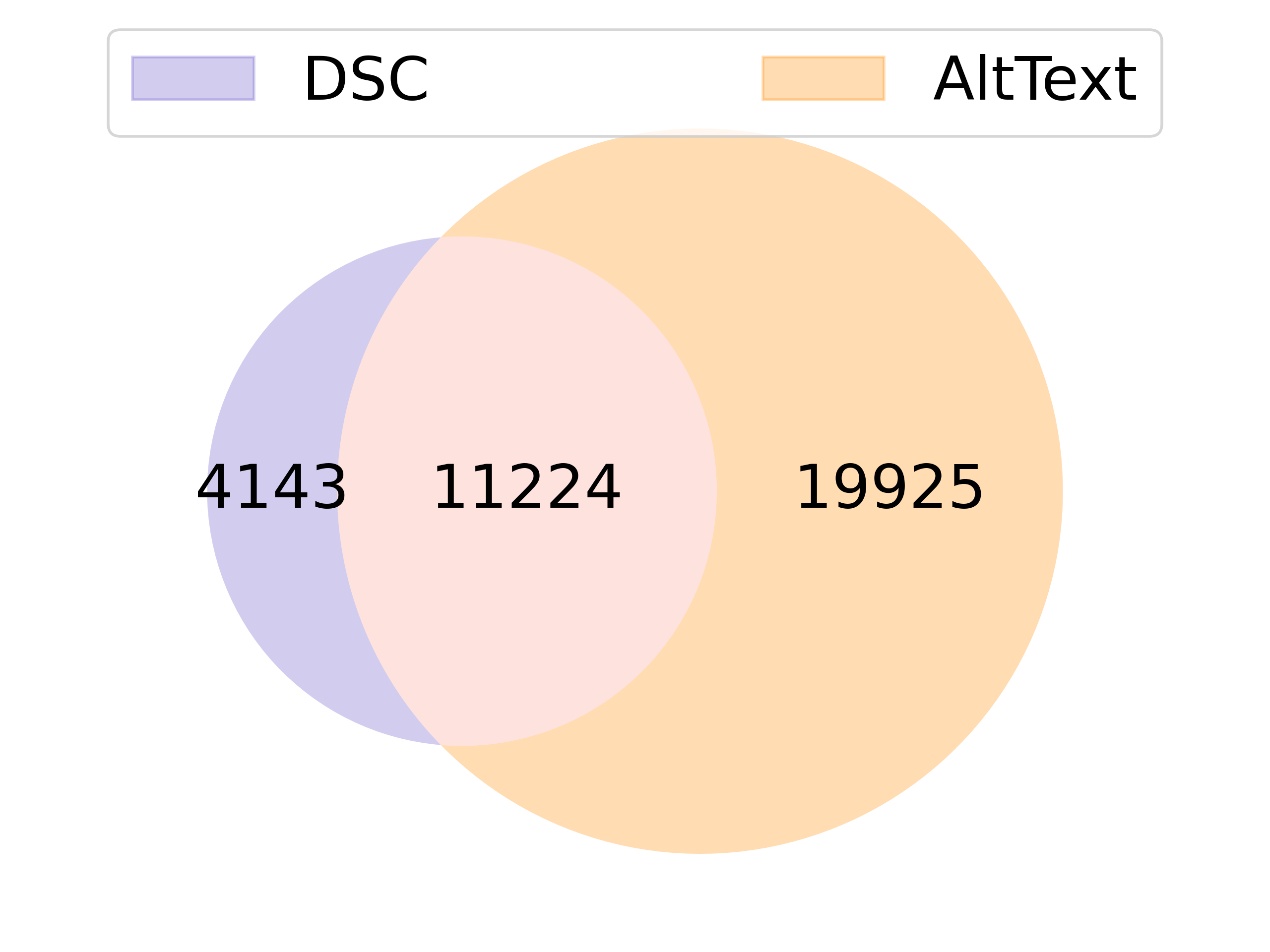} \hspace{-8mm}
 
	 & \includegraphics[width=0.27\textwidth]{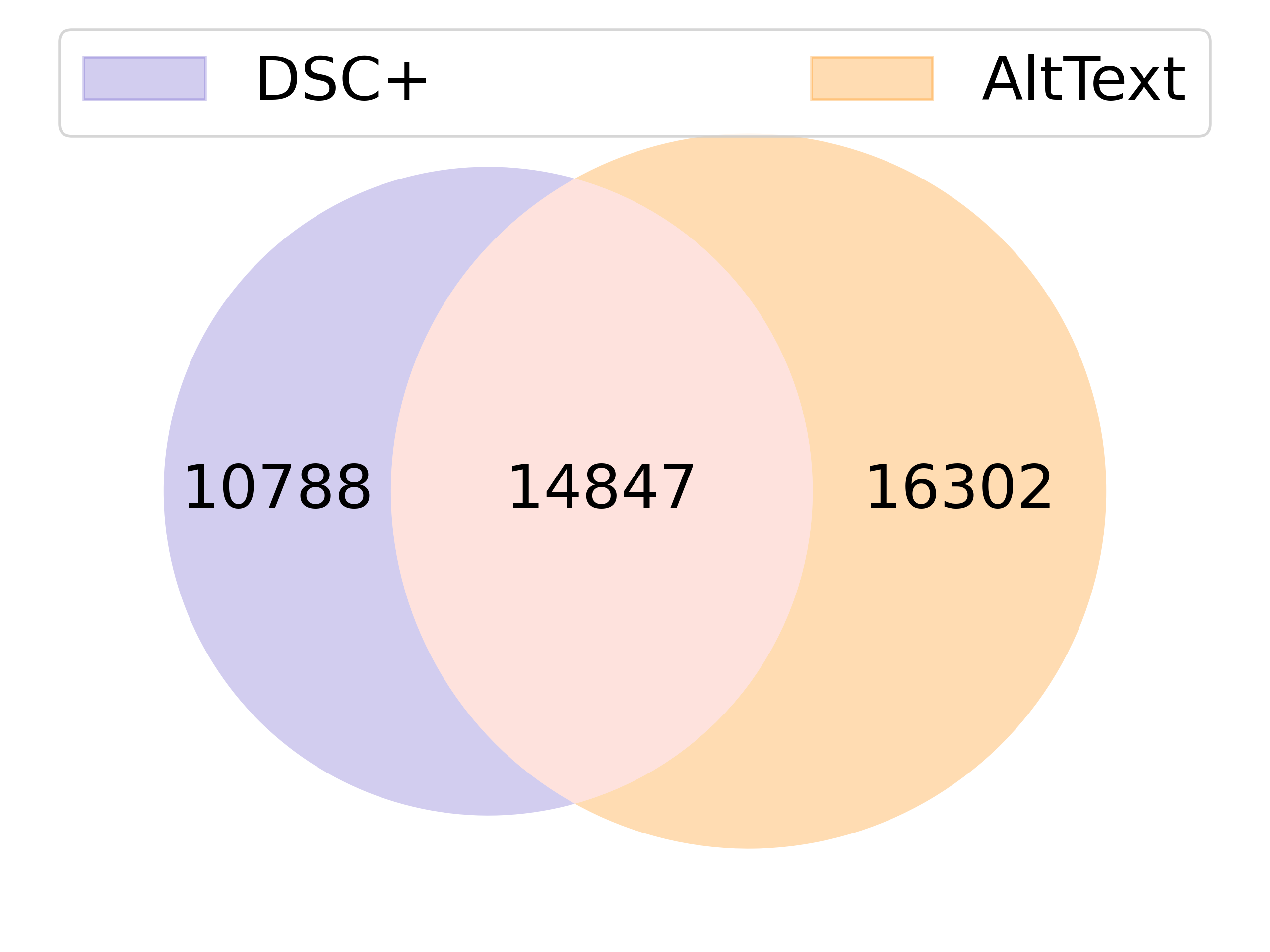} \hspace{-8mm} &
	\includegraphics[width=0.27\textwidth]{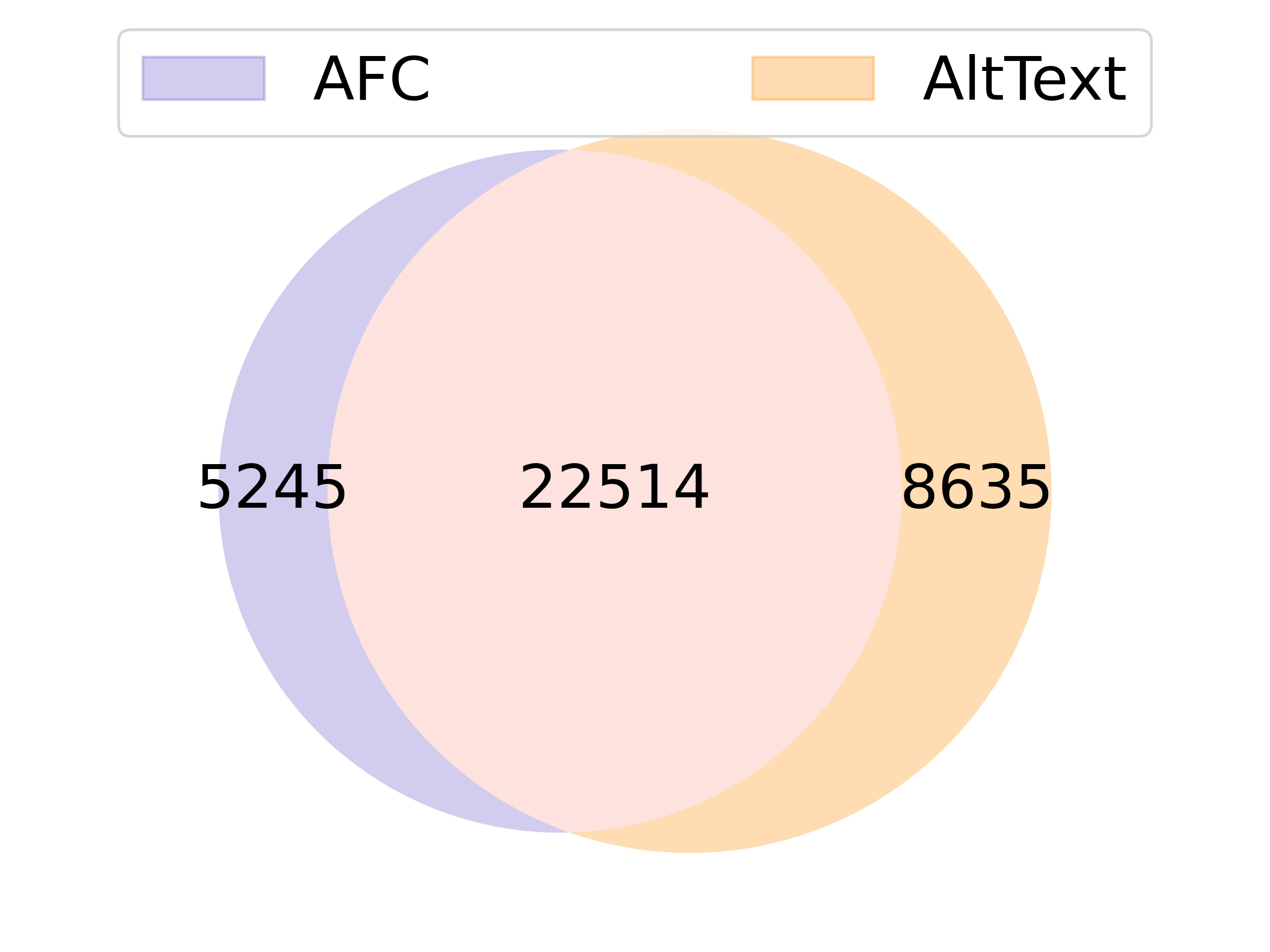}  \\
	\end{tabular}
\end{center}
\vspace{-0.5cm}
\caption{The number of unique entities in different synthetic captions (randomly sample 17.5k images) compared to AltText: AltText provides more unique entities as wider knowledge.}
\label{fig:entity}
\end{figure*}
\textbf{Richness assessment: token length and average number of assertions.}
Fig.~\ref{fig:captions_lengths} illustrates the distribution of the number of tokens of various caption types generated in this study. Specifically, SSC mainly ranges from 10 to 15 tokens, while DSC spans from 40 to 60 tokens, both fitting within the text encoder's capacity in CLIP. In contrast, most DSC+ captions exceed 100 tokens. Besides that, we propose Average Number of Assertions (ANA) to quantify the richness of captions. We prompt an LLM to generate different assertions of a caption to analyze our different formats of captions in terms of the richness. More details of this approach is in Appendix. Note that the ANA for SSC is 2.49, DSC as 8.13 and DSC+ as 12.20, showing more visual contents.

\textbf{Diversity assessment: number of unique entities in captions.} 
We hypothesize that the original, albeit noisy, AltText may carry a broader range of diverse information and knowledge, offering potential advantages for CLIP’s pre-training. To assess this diversity, we quantify the number of unique entities present in the captions and present a visualization in Fig.~\ref{fig:entity}. Our analysis shows that AltText contains a higher number of unique entities, which could be beneficial in providing a wider knowledge base.

\vspace{-0.2cm}

\section{Image-Caption Data for Multimodal Foundation Models}
In this section, we mainly discuss three foundation models: CLIP, multimodal LLM, and diffusion models. For both CLIP and diffusion models, since the text encoder is limited to 77 tokens~\citep{zhang2024longclip}, we focus primarily on SSC and DSC. For multimodal LLM, we explore more detailed versions, including DSC+ and AFC. We summarize our key findings below:
\begin{itemize}[leftmargin=*]
    \item The tradeoff between the richness of captions and their accuracy needs to be balanced based on the multimodal tasks.  
    \item Both AltText and synthetic captions are important for CLIP training, with shorter captions yielding better performance. Linear probing is an additional effective way to evaluate the representations. 
    \item Pre-training and SFT benchmarks can behave differently in multimodal LLMs. On the SFT benchmarks, MM1 shows a preference for DSC+ alone.
    \item For diffusion models, DSC emerges as the most effective captioning strategy.
\end{itemize}

\subsection{Image-Caption Data for CLIP}

We use VeCap-300M~\citep{veclip}, a web-crawled dataset with raw AltText as our main pre-training dataset for CLIP. Besides AltText, we generate several synthetic caption datasets for the study. Then, we use ViT-B/16 as the vision encoder. The training details can be found in Appendix.

\begin{table}[t]
\begin{tabular}{cc}

\hspace{-0.15cm}%
\begin{minipage}[t]{0.57\linewidth}
\vspace{0pt}
\centering

\caption{\small{Effect of different synthetic captions on CLIP with ViT-B/16 as the backbone. IN: ImageNet.
}}
\vspace{-0.2cm}
\label{tab:vecap_clip}   
\small
\resizebox{1.0\textwidth}{!}{
  \begin{tabular}{ccccc|cc}
        \toprule[1.2pt]
        \multirow{2}{*}{\bf Pre-train Caption}  & \multicolumn{2}{c}{\bf COCO (R@1)} & \multicolumn{2}{c}{\bf Flickr30k (R@1)} & \multirow{2}{*}{\bf IN} & \multirow{2}{*}{\bf INV2} \\
         & \bf I-T & \bf T-I & \bf I-T & \bf T-I & & \\
        \midrule
        AltText  & 54.24 & 36.98 & 81.30 & 65.80 & 65.70 & 58.58 \\
        DSC  & 52.28 & 29.00 & 80.90 & 54.75 & 27.30 & 21.91 \\
        SSC & 57.00 & 35.15 & 84.67 & 63.35 & 49.10 & 43.31 \\
        AFC &54.82 &	34.84 &	84.00&	62.18&	38.98	&35.11 \\
        DSC + AltText  &65.84	&46.08	&90.26	&73.94	&66.18	&58.74 \\
        SSC + AltText  & 66.67	&48.13	&91.81	&76.54	&\textbf{66.63}	& \textbf{59.57} \\
        AFC + AltText &63.98	&43.76	&89.10	&73.32	&66.47	&58.84 \\
        All Synthetic + AltText  &\textbf{70.12}	&\textbf{50.21}	& \textbf{93.00}	& \textbf{77.72}	&64.91	&57.92 \\


        \bottomrule[1.2pt]
    \end{tabular}
}

\end{minipage}

&
\hspace{-0.3cm}%
\begin{minipage}[t]{0.4\linewidth}
\vspace{0pt}


\centering

\caption{\small{Evaluation with linear probing (LP) on ImageNet for CLIP.
}}
\vspace{-0.2cm}
\label{tab:linear_probing}   
\small
\resizebox{1.0\textwidth}{!}{
  \begin{tabular}{c|ccc}
        \toprule[1.2pt]
        \bf Pre-train Caption & \bf Zero-shot & \bf LP & \bf Gain \\
        \midrule
        AltText  & 65.70 &	78.34	& +12.64 \\
        DSC & 27.30	& 75.72	&  +48.42 \\
        SSC & 66.63	& 79.94 & 	+13.31 \\
        AFC & 38.98	& 75.53 &	+36.55 \\
        DSC + AltText & 66.18	& 79.96	 & +13.78 \\
        SSC + AltText & 66.63	& 79.94	& +13.31 \\
        AFC + AltText & 66.47	& 78.70 & +12.23 \\
        DSC + SSC + AltText & 65.16 & 80.01 & +14.85 \\
        All Synthetic + AltText  &64.91	 &79.55	 & +14.64 \\

        \bottomrule[1.2pt]
    \end{tabular}
}
\end{minipage}

\end{tabular}
\vspace{-0.1cm}
\end{table}

\textbf{Effect of synthetic captions.} We first study the effect of Short Synthetic Captions (SSC) and Descriptive Synthetic Captions (DSC).
Results are summarized in Table~\ref{tab:vecap_clip}.
Interestingly, while SSC enhances retrieval performance, it leads to a substantial drop in zero-shot ImageNet accuracy. Moreover, despite demonstrating that synthetic captions have superior image-text alignment and reduced noise, the more descriptive captions (DSC) perform worse across all benchmarks.   Based on these results, DSC appears suboptimal for CLIP training, leading to inferior performance. For example, despite DSC capturing more visual concepts within the captions, its zero-shot performance shows a significant degradation of 21.8\% compared to SSC. We hypothesize this performance drop may be partially due to a distribution mismatch as the prompts in COCO/Flickr30k/ImageNet datasets are short (e.g., ``a photo of \{\}'').

To further explore the performance gap between DSC and SSC on CLIP, we also use linear probing, which provides a direct measure of the quality and generalization capability of the representations learned by CLIP. Strong performance from a linear classifier on specific tasks indicates that the pre-trained model has effectively captured relevant and discriminative features, underscoring the robustness of its embeddings. 
We summarize the results on linear probing in Table~\ref{tab:linear_probing}. Even though DSC and SSC show lower zero-shot performance, they achieve comparable results to AltText after linear probing, indicating similar pre-trained representations. 
This indicates that the relatively poor zero-shot results of DSC can be significantly improved with linear probing, implying that representations learned from DSC are richer than initially presumed. 

\textbf{Intersection of synthetic captions and AltText.} From Table~\ref{tab:vecap_clip}, we find training CLIP solely on large-scale synthetic captions may get inferior results compared to the original AltText, even though synthetic captions have better image-text alignment. We hypothesize that synthetic captions rewritten from a MLLM may hurt the diversity and knowledge coverage of the original AltText. Therefore, we blend our synthetic captions and AltText, such as DSC + AltText and SSC + AltText: there is a significant boost in retrieval performance—e.g., over 10\% improvement on COCO. Additionally, SSC + AltText also leads to gains in ImageNet accuracy.  Utilizing a mixture of all synthetic caption formats  yields the best performance in retrieval tasks, likely due to the wider entity-based knowledge, as visualized in Fig.~\ref{fig:entity}.

\begin{figure}[t]
    \centering
    \includegraphics[width=1.0\linewidth]{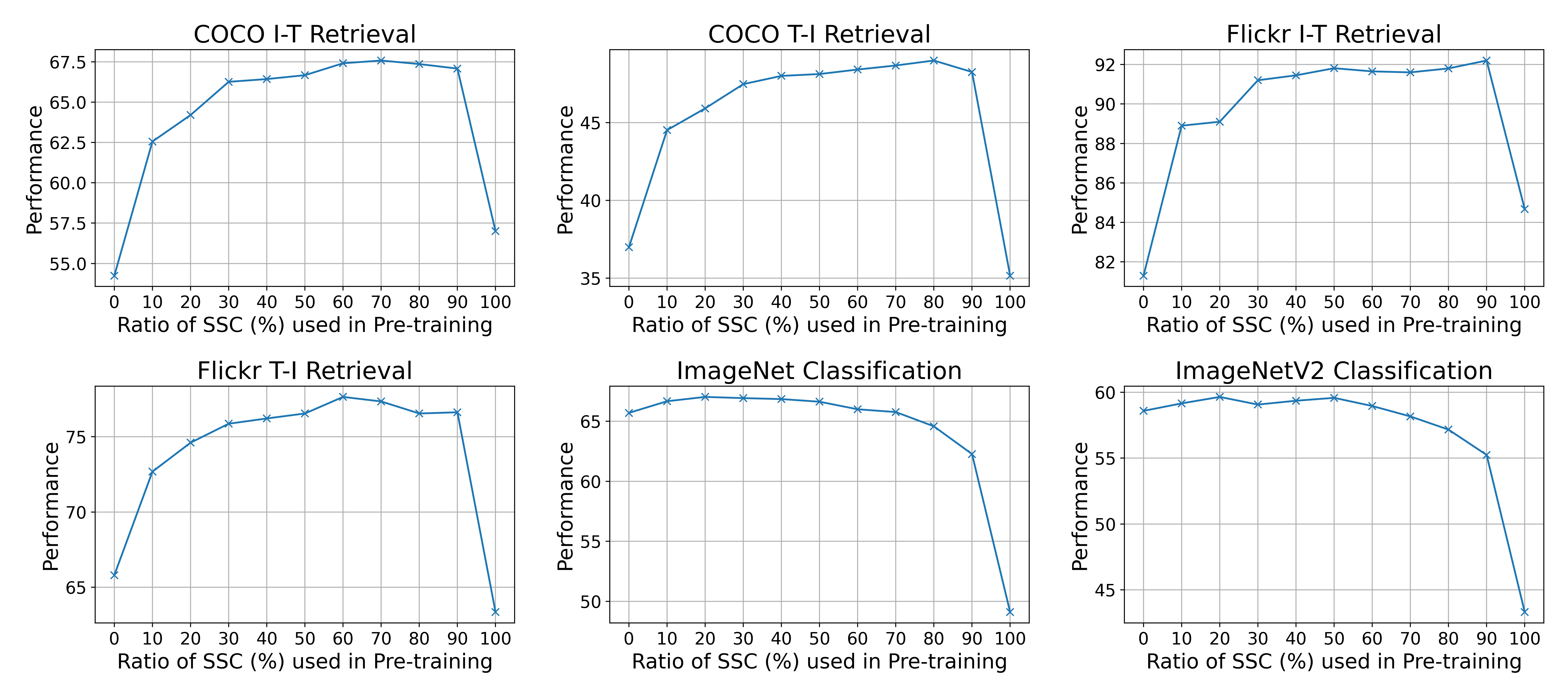}
    \vspace{-0.5cm}
    \caption{The intersection of synthetic captions and AltText for CLIP. We gradually increase the proportion of SSC mixed with AltText during training.  All experiments use ViT-B/16 as the backbone and VeCap-300M~\citep{veclip} as the pre-training dataset.}
    \label{fig:clip_ratio_study}
    \vspace{-0.2cm}
\end{figure} 
\textbf{Optimal mixture ratio between synthetic captions and AltText. }
Considering the wider knowledge of AltText (Fig.~\ref{fig:entity}) and the better alignment of synthetic captions, we explore the optimal mixture ratio. 
Specficially, we use SSC and AltText from VeCap-300M~\citep{veclip} as an example, with results shown in Fig.~\ref{fig:clip_ratio_study}. A ratio of 0 corresponds to using only SSC, while 100 corresponds to using only AltText. We observe that CLIP achieves optimal performance across both retrieval and classification tasks when the ratio is tuned to around 40-50\%. A lower proportion of AltText leads to a drop in retrieval performance, whereas a lower proportion of SSC results in decreased accuracy in ImageNet zero-shot classification. 
This observation aligns with findings in~\citet{li2024if}. From this, we verify that AltText provides broader knowledge coverage and greater diversity, which benefits CLIP's pre-training by enabling the model to grasp a wider range of concepts. This diversity may serve as a foundation for generalization, allowing CLIP to better represent varied contexts and domains in zero-shot classification.

\textbf{Exploration of the optimal way of using AltText: AFC vs. simple mixture.} As shown in Fig.~\ref{fig:clip_ratio_study} and Table~\ref{tab:vecap_clip}, we verify that a simple mixture of our synthetic captions and AltText can achieve superior results. 
In addition to this straightforward combination, we investigate alternative methods of incorporating AltText into the training process. One promising approach is to fuse the knowledge from AltText directly into the synthetic captions. To enable this, we fine-tune our captioner with instruction-following capabilities, generating AltText Fusion Captions (AFC). While AFC shows improvement over DSC due to the enriched AltText information, it underperforms when compared to the simple mixture of DSC and AltText.

\subsection{Image-Caption Data for Multimodal LLM}
\begin{table}[ht]
\begin{tabular}{cc}

\hspace{-0.15cm}%
\begin{minipage}[t]{0.48\linewidth}
\vspace{0pt}
\centering

\caption{\small{The effect of synthetic captions on MM1 (1.2B model) pre-training. 
}}
\vspace{-0.2cm}
\label{tab:mm1}   
\small
\resizebox{1.0\textwidth}{!}{
  \begin{tabular}{c|cccc}
        \toprule[1.2pt]
        Pre-train Caption & TextCore & 0-Shot & 4-Shot & 8-Shot \\
        \midrule
        AltText & 53.48 & 34.80 & 55.81 & 59.72 \\
        DSC & \textbf{53.76} & 37.05 & 59.36 & 63.68 \\
        DSC + AltText & 53.71 & 37.09 & \textbf{60.31} & \textbf{63.96} \\
        SSC & 53.46 & \textbf{37.35} & 59.19 & 63.56 \\ 
        SSC + AltText & 53.20 & 37.16 & 58.22 & 62.22 \\

        \bottomrule[1.2pt]
    \end{tabular}
}

\end{minipage}

&
\hspace{-0.1cm}%
\begin{minipage}[t]{0.45\linewidth}

\vspace{0pt}
\centering

\caption{\small{The ratio ablation on MM1  pre-training between synthetic captions and AltText. 
}}
\vspace{-0.2cm}
\label{tab:mm1_ratio}   
\small
\resizebox{1.0\textwidth}{!}{
  \begin{tabular}{c|cccc}
        \toprule[1.2pt]
        Mixing Ratio & TextCore & 0-Shot & 4-Shot & 8-Shot \\
        \midrule
        33/66 & 53.86 & 35.16 & 59.69 & 63.97 \\
        50/50 & 53.86 & 36.40 & 60.17 & 64.13 \\
        66/33 & \textbf{54.24} & \textbf{37.02} & \textbf{60.26} & \textbf{64.71}  \\
        80/20 & 53.96 & 35.74 & 60.09 & 64.69 \\
        100/0 & 54.19 & 15.54 & 60.24 & 64.02 \\

        \bottomrule[1.2pt]
    \end{tabular}
}

\end{minipage}

\end{tabular}
\vspace{-0.1cm}
\end{table}

As we study large-scale image-caption data for MLLMs, we use MM1~\citep{mm1} as one example and focus on the pre-training stage. MM1~\citep{mm1} claims that captioning data lift the zero-shot performance and synthetic captions are helpful for few-shot learning. Based on this insight, we further study the captioning data recipe on how to balance the use of original AltText and synthetic captions. 
We follow the pre-training setup and the evaluation benchmark in MM1~\citep{mm1} to report TextCore and 0/4/8-shot performance.
All of the experiments are conducted on the 1.2B model. We pre-train the model with 50K steps and the batch size is 512. More details are in Appendix.

\begin{table}[t]
\centering
\caption{\small SFT evaluation of models pre-trained with different types of captions. We use the same SFT evaluation benchmarks as in MM1~\citep{mm1}. (*): Concatenation of two captions.}
    \vspace{-0.2cm}
    \small
    \resizebox{\textwidth}{!}{
\begin{tabular}{c|ccccccccc|c}
\toprule[1.2pt]
\textbf{Pre-Trained Data} & \textbf{VQAv2} & \textbf{VQA$^\text{T}$} & \textbf{MMMU} & \textbf{MathV} & \textbf{MME$^\text{P}$} & \textbf{MME$^\text{C}$} & \textbf{SEED} & \textbf{POPE} & \textbf{LLaVA$^\text{W}$} & \textbf{Average} \\
\midrule 

AltText & 77.1 & 66.1 & 31.4 & 28.3 & 781.7 & 225.4 & 61.7 & 84.6 & 69.8 & 57.6 \\
LLaVA Caption & 78.0 & 65.4 & 30.0 & 27.6 & 773.4 & 200.4 & 63.5 & 83.7 & 63.6 & 56.5 \\
SSC & 78.8 & 65.8 & \textbf{33.6} & 28.2 & 760.2 & 216.4 & 63.1 & 84.4 & 69.0 & 57.9 \\
DSC & 77.1 & 61.8 & 30.8 & 27.9 & 596.6 & 213.6 & 64.6 & 83.9 & 70.8 & 56.3 \\
DSC+ & 79.0 & 65.4 & 32.6 & 29.4 & 727.2 & 224.3 & \textbf{66.8} & 85.1 & 71.5 & \textbf{58.7} \\
AltText + SSC (*) & 77.7 & 65.4 & 32.3 & 27.6 & \textbf{781.9} & \textbf{236.1} & 62.0 & 84.2 & 70.3 & 57.7 \\
AltText + DSC (*) & 78.2 & 66.9 & 31.9 & \textbf{30.7} & 686.4 & 216.4 & 63.8 & 84.3 & \textbf{72.0} & 58.2 \\
AltText + DSC+ (*) & 79.2 & \textbf{66.9} & 31.1 & 29.6 & 740.2 & 229.3 & 65.5 & 85.1 & 68.1 & 58.2 \\
AFC & 78.0 & 65.6 & 32.3 & 29.4 & 713.2 & 214.3 & 66.0 & 84.4 & 70.0 & 58.0 \\
SSC + DSC + DSC+ & \textbf{79.4} & 66.5 & 30.2 & 30.3 & 689.3 & 198.9 & 65.5 & 83.8 & 67.5 & 57.5 \\
AFC + SSC + DSC + DSC+ & 77.3 & 62.6 & 31.6 & 27.7 & 661.5 & 198.2 & 64.1 & 83.5 & 64.2 & 55.9 \\
 \bottomrule[1.2pt] 
 
 \end{tabular}
}
\label{tab:mm1_all_sft_results}
    \vspace{-0.3cm} 
\end{table}


\textbf{Effect of synthetic captions for pre-training benchmarks.} 
We generate DSC and SSC captions for VeCap-300M~\citep{veclip}, as used in MM1~\citep{mm1}, and replace the captions during MM1 pre-training. The results, summarized in Table~\ref{tab:mm1}, show that our synthetic captions yield improved performance in image-text benchmarks across 0-shot to 8-shot settings. For example, SSC achieves a \textbf{+1.3\%} performance gain in 0-shot evaluation compared to the original MM1. Unlike CLIP experiments, DSC outperforms SSC, with the combination of DSC and original AltText delivering the best results in this context. We also conduct an ablation study of data mixing ratios on MM1 pre-training to explore the optimal balance between synthetic captions and AltText, as summarized in Table~\ref{tab:mm1_ratio}. The results indicate that a 66/33 mixing ratio yields the best overall performance across all evaluation settings. Specifically, this ratio achieves the highest scores for TextCore (54.24), 0-shot (37.02), 4-shot (60.26), and 8-shot (64.71) performance. While increasing the proportion of synthetic captions generally improves performance, there is a significant drop in 0-shot performance when using only synthetic captions (100/0 ratio).

\textbf{Effect of synthetic captions for SFT benchmarks.} 
Besides pre-training benchmarks, we also conduct SFT and then evaluate the model to analyze the profound effect of image-caption data in MLLMs. We use the same SFT recipe to have a fair comparison. As shown in Table~\ref{tab:mm1_all_sft_results}, DSC+ and the concatenation of AltText with DSC+ deliver the best performance. 
This strongly suggests the importance of detailed captions, despite them containing potentially the highest number of hallucinations among all the caption types we test. On the other hand, concatenating AltText with synthetic captions does not yield significant improvement in the SFT benchmarks, contrasting with the gains observed in pre-training benchmarks. We hypothesize that the primary role of image-caption data during the pre-training phase of MLLMs is to enhance image-text alignment. Consequently, more detailed captions, such as DSC+, deliver superior results after the SFT stage.

\textbf{DSC+ alone can outperform diverse synthetic captions.}  Unlike CLIP, where mixing diverse synthetic captions leads to superior results, for MM1, detailed captions (DSC+) alone yield the best performance after the SFT stage. 
As shown in Table~\ref{tab:mm1_all_sft_results}, DSC+ achieves a 58.7\% score, outperforming LLaVA captions (56.5\%) by 2.2\%. This suggests that providing richer and more specific information in captions helps multimodal LLMs like MM1 generalize better after the SFT stage. The combination of SSD, DSC, DSC+, and AFC does not lead to better results, suggesting that multimodal LLMs may benefit more from detailed captions during pre-training. These findings suggest that while combining synthetic captions proves beneficial in some contexts (e.g., CLIP), for multimodal LLMs like MM1, a single, detailed caption offers more effective guidance during pre-training.

\subsection{Image-Caption Data for Diffusion Model}
Inspired by DALLE-3~\citep{dalle3}, detailed and short captions can improve the prompt following ability. In this work, our DSC not only covers the main  objects  within the scene, but also their relationships, attributes, and the broader context in which they are situated. We hypothesize that this level of detail allows the model to generate images that are not only visually accurate but also semantically aligned with the textual input. We implement Stable Diffusion 3~\citep{sd3} and use this diffusion model as our studying example on text-to-image generation tasks. The backbone is based on the DiT architecture~\citep{dit} that focuses exclusively on class-conditional image generation and incorporates a modulation mechanism to condition the network based on both the diffusion process timestep and the class label. Different from DALLE-3~\citep{dalle3}, we report results on more comprehensive benchmarks instead of only CLIP score, such as GenEval~\citep{geneval} and DSG~\citep{DSG}.

\textbf{Effect of synthetic captions.}
Synthetic captions lead to significant improvements on the GenEval benchmark~\citep{geneval}, as shown in Table~\ref{tab:diffusion}, highlighting the advantage of enhanced prompt-following capabilities. Notably, incorporating SSC or DSC with AltText boosts the GenEval average score from 58.8 to 65.5. Additionally, synthetic captions yield over a 3.5\% improvement on the DSG benchmark~\citep{DSG}. However, SSC achieves better performance on the FID score~\citep{FID}. Overall, the descriptive captions show better results among these benchmarks.

\begin{figure}[t]
    \centering
    \includegraphics[width=1.0\linewidth]{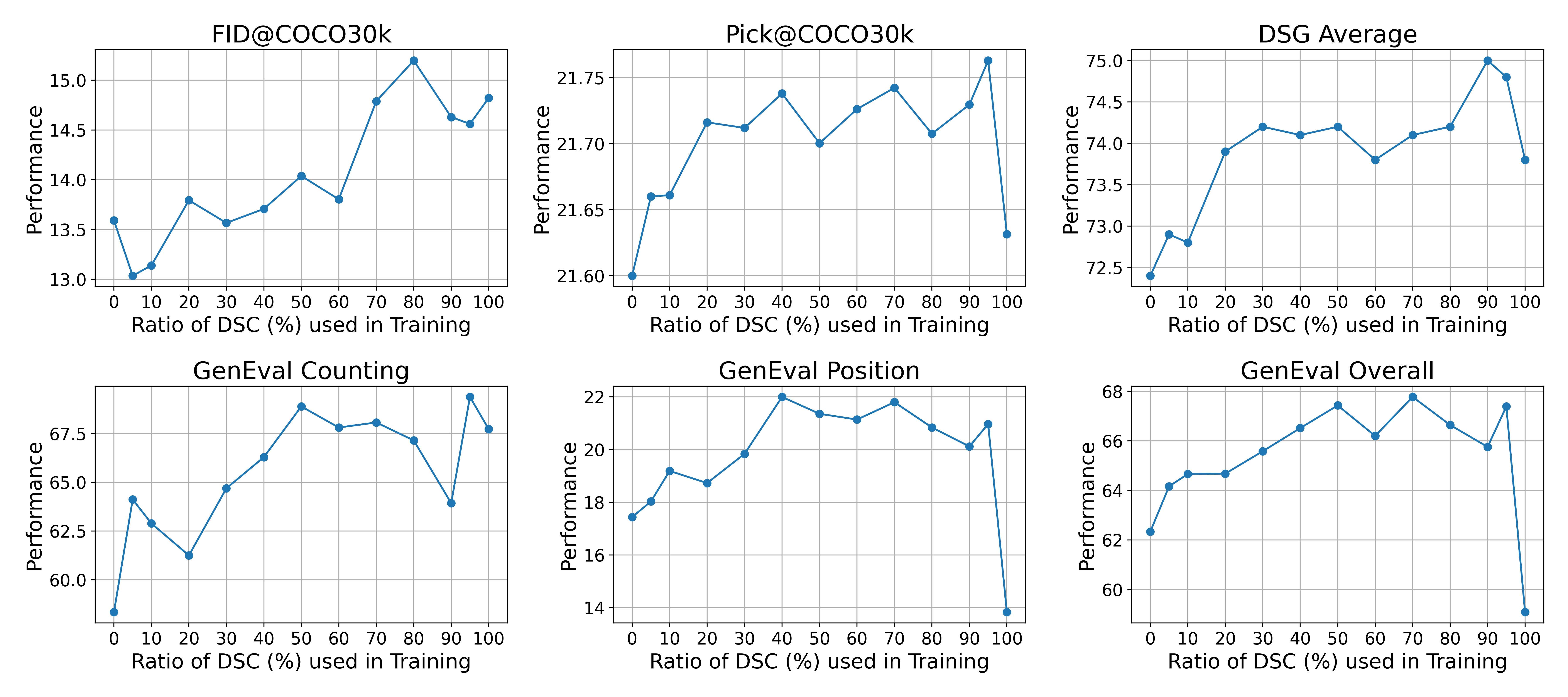}
    \vspace{-0.5cm}
    \caption{The intersection of synthetic captions and AltText for  diffusion models. We gradually increase the proportion of DSC mixed with AltText during training.}
    \label{fig:diffusion_ratio_study}
    \vspace{-0.2cm}
\end{figure}

\textbf{Ablation study on mixing ratio of synthetic captions and AltText.} We examine the impact of varying the ratio between DSC and AltText in diffusion model training, evaluating performance across FID@COCO30k~\citep{FID}, CLIP@COCO30k, DSG Average~\citep{DSG}, GenEval Overall~\citep{geneval}. Results are summarized in Fig.~\ref{fig:diffusion_ratio_study}. The FID@COCO30k metric shows a gradual increase, suggesting that higher DSC ratios lead to improvements in generation quality. The DSG Average score exhibits improvements with a higher DSC ratio, indicating that DSC can enhance the model's ability to handle complex tasks. However, the performance on the GenEval related metric peaks at 50\% DSC, after which it begins to decline, highlighting the necessity of balancing synthetic and original captions to achieve optimal results across diverse evaluation merics. Besides DSC, we further conduct experiments using SSC as well. Results are summarized in Table \ref{tab:diffusion}. Overall, the use of SSC alsone also achieves competitive performance, but the use of DSC and AltText together appears to be a better captioning strategy.

\begin{table*}[t!]
    \centering
    \caption{\small The effect of synthetic captions on diffusion models.}
    \vspace{-0.2cm}
    \small
    \resizebox{0.95\textwidth}{!}{
    \begin{tabular}{c|ccccccc|c|c}
    \toprule[1.2pt]
\multirow{2}{*}{} & \multicolumn{7}{c|}{GenEval~\citep{geneval}} & \multirow{2}{*}{\begin{tabular}[c]{@{}c@{}}FID \\ COCO30k\end{tabular}} & \multirow{2}{*}{\begin{tabular}[c]{@{}c@{}}DSG\\ Average\end{tabular}} \\
& Single Obj & Two Obj & Counting & Colors & Position & Attribution & Average & & \\
    \midrule
AltText & 99.1 & 80.4 & 58.3 & 78.5 & 17.4 & 40.2 & 62.3 & 13.6 & 72.4 \\ 
SSC & 98.5 & 80.1 & 68.9 & 80.4 & 18.6 & 50.2 & 66.1 & 13.1 & 73.1\\
SSC + AltText & 98.2 & 84.4 & 65.1 & 77.3 & 18.6 & 50.2 & 65.5 & 13.1 & 74.2 \\
DSC & 93.9 & 60.2 & 67.7 & 79.8 & 13.8 & 39.2 & 59.1 & 14.8 & 73.8 \\
DSC + AltText & 99.2 & 84.2 & 68.9 & 79.0 & 21.4 & 52.0 & 67.4 & 14.0 & 74.2 \\

 \bottomrule[1.2pt]                                                               
\end{tabular}
    }
    \label{tab:diffusion}
    \vspace{-0.3cm}
\end{table*}

\section{Discussion}
In this study, we examine the role and value of image-text data in multimodal foundation models, including CLIP, multimodal LLMs, and diffusion models. Our research focuses on the intersection between synthetic image-aligned captions and the original web-crawled AltText. To identify the most effective captions for each foundation model, we develop a controllable and human-aligned captioning pipeline designed to minimize hallucinations and generate various types of captions as needed. Through extensive pre-training experiments, we derive the following key insights. 1) Both AltText and synthetic captions play crucial roles—AltText contributes to more diverse information, while synthetic captions offer improved image-text alignment. 2) CLIP tends to favor short synthetic captions, whereas  MLLMs benefit from more descriptive captions. We also observe that the benchmarks in the pre-training and SFT stage of MLLMs may have different preferences of captions. 3) We verify the observation from DALLE-3 on text-to-image generation with more comprehensive benchmarks and show the benefits of synthetic captions. 
For future work, we aim to further refine our captioning pipeline, enhancing its ability to generate task-specific captions across a wider range of multimodal applications.

\bibliography{iclr2025_conference}

\begin{thebibliography}{54}
\providecommand{\natexlab}[1]{#1}
\providecommand{\url}[1]{\texttt{#1}}
\expandafter\ifx\csname urlstyle\endcsname\relax
  \providecommand{\doi}[1]{doi: #1}\else
  \providecommand{\doi}{doi: \begingroup \urlstyle{rm}\Url}\fi

\bibitem[Berant et~al.(2013)Berant, Chou, Frostig, and Liang]{berant2013semantic}
Jonathan Berant, Andrew Chou, Roy Frostig, and Percy Liang.
\newblock Semantic parsing on freebase from question-answer pairs.
\newblock In \emph{Proceedings of the 2013 conference on empirical methods in natural language processing}, pp.\  1533--1544, 2013.

\bibitem[Betker et~al.(2023)Betker, Goh, Jing, Brooks, Wang, Li, Ouyang, Zhuang, Lee, Guo, et~al.]{dalle3}
James Betker, Gabriel Goh, Li~Jing, Tim Brooks, Jianfeng Wang, Linjie Li, Long Ouyang, Juntang Zhuang, Joyce Lee, Yufei Guo, et~al.
\newblock Improving image generation with better captions.
\newblock \emph{Computer Science. https://cdn. openai. com/papers/dall-e-3. pdf}, 2\penalty0 (3):\penalty0 8, 2023.

\bibitem[Bisk et~al.(2020)Bisk, Zellers, Gao, Choi, et~al.]{bisk2020piqa}
Yonatan Bisk, Rowan Zellers, Jianfeng Gao, Yejin Choi, et~al.
\newblock Piqa: Reasoning about physical commonsense in natural language.
\newblock In \emph{Proceedings of the AAAI conference on artificial intelligence}, volume~34, pp.\  7432--7439, 2020.

\bibitem[Bradbury et~al.(2018)Bradbury, Frostig, Hawkins, Johnson, Leary, Maclaurin, Necula, Paszke, Vander{P}las, Wanderman-{M}ilne, and Zhang]{jax2018github}
James Bradbury, Roy Frostig, Peter Hawkins, Matthew~James Johnson, Chris Leary, Dougal Maclaurin, George Necula, Adam Paszke, Jake Vander{P}las, Skye Wanderman-{M}ilne, and Qiao Zhang.
\newblock {JAX}: composable transformations of {P}ython+{N}um{P}y programs.
\newblock \emph{Github}, 2018.
\newblock URL \url{http://github.com/google/jax}.

\bibitem[Chen et~al.(2023)Chen, Zhang, Zeng, Zhang, Zhu, and Zhao]{shikra}
Keqin Chen, Zhao Zhang, Weili Zeng, Richong Zhang, Feng Zhu, and Rui Zhao.
\newblock Shikra: Unleashing multimodal llm's referential dialogue magic.
\newblock \emph{arXiv preprint arXiv:2306.15195}, 2023.

\bibitem[Chen et~al.(2024{\natexlab{a}})Chen, Li, Dong, Zhang, He, Wang, Zhao, and Lin]{sharegpt4v}
Lin Chen, Jisong Li, Xiaoyi Dong, Pan Zhang, Conghui He, Jiaqi Wang, Feng Zhao, and Dahua Lin.
\newblock Sharegpt4v: Improving large multi-modal models with better captions.
\newblock \emph{ECCV}, 2024{\natexlab{a}}.

\bibitem[Chen et~al.(2024{\natexlab{b}})Chen, Wu, Wang, Su, Chen, Xing, Zhong, Zhang, Zhu, Lu, et~al.]{internvl}
Zhe Chen, Jiannan Wu, Wenhai Wang, Weijie Su, Guo Chen, Sen Xing, Muyan Zhong, Qinglong Zhang, Xizhou Zhu, Lewei Lu, et~al.
\newblock Internvl: Scaling up vision foundation models and aligning for generic visual-linguistic tasks.
\newblock In \emph{Proceedings of the IEEE/CVF Conference on Computer Vision and Pattern Recognition}, pp.\  24185--24198, 2024{\natexlab{b}}.

\bibitem[Cho et~al.(2024)Cho, Hu, Baldridge, Garg, Anderson, Krishna, Bansal, Pont-Tuset, and Wang]{DSG}
Jaemin Cho, Yushi Hu, Jason Baldridge, Roopal Garg, Peter Anderson, Ranjay Krishna, Mohit Bansal, Jordi Pont-Tuset, and Su~Wang.
\newblock Davidsonian scene graph: Improving reliability in fine-grained evaluation for text-to-image generation.
\newblock In \emph{ICLR}, 2024.

\bibitem[Clark et~al.(2018)Clark, Cowhey, Etzioni, Khot, Sabharwal, Schoenick, and Tafjord]{ARC}
Peter Clark, Isaac Cowhey, Oren Etzioni, Tushar Khot, Ashish Sabharwal, Carissa Schoenick, and Oyvind Tafjord.
\newblock Think you have solved question answering? try arc, the ai2 reasoning challenge.
\newblock \emph{arXiv preprint arXiv:1803.05457}, 2018.

\bibitem[Dai et~al.(2023)Dai, Li, Li, Tiong, Zhao, Wang, Li, Fung, and Hoi]{instructblip}
Wenliang Dai, Junnan Li, Dongxu Li, Anthony Meng~Huat Tiong, Junqi Zhao, Weisheng Wang, Boyang Li, Pascale Fung, and Steven Hoi.
\newblock Instructblip: Towards general-purpose vision-language models with instruction tuning, 2023.

\bibitem[Esser et~al.(2024)Esser, Kulal, Blattmann, Entezari, M{\"u}ller, Saini, Levi, Lorenz, Sauer, Boesel, et~al.]{sd3}
Patrick Esser, Sumith Kulal, Andreas Blattmann, Rahim Entezari, Jonas M{\"u}ller, Harry Saini, Yam Levi, Dominik Lorenz, Axel Sauer, Frederic Boesel, et~al.
\newblock Scaling rectified flow transformers for high-resolution image synthesis.
\newblock In \emph{Forty-first International Conference on Machine Learning}, 2024.

\bibitem[Fan et~al.(2024)Fan, Krishnan, Isola, Katabi, and Tian]{laclip}
Lijie Fan, Dilip Krishnan, Phillip Isola, Dina Katabi, and Yonglong Tian.
\newblock Improving clip training with language rewrites.
\newblock \emph{Advances in Neural Information Processing Systems}, 36, 2024.

\bibitem[Fang et~al.(2023)Fang, Jose, Jain, Schmidt, Toshev, and Shankar]{DFN}
Alex Fang, Albin~Madappally Jose, Amit Jain, Ludwig Schmidt, Alexander~T Toshev, and Vaishaal Shankar.
\newblock Data filtering networks.
\newblock In \emph{The Twelfth International Conference on Learning Representations}, 2023.

\bibitem[Fu et~al.(2024)Fu, Chen, Shen, Qin, Zhang, Lin, Yang, Zheng, Li, Sun, Wu, and Ji]{mme}
Chaoyou Fu, Peixian Chen, Yunhang Shen, Yulei Qin, Mengdan Zhang, Xu~Lin, Jinrui Yang, Xiawu Zheng, Ke~Li, Xing Sun, Yunsheng Wu, and Rongrong Ji.
\newblock Mme: A comprehensive evaluation benchmark for multimodal large language models, 2024.

\bibitem[Gadre et~al.(2024)Gadre, Ilharco, Fang, Hayase, Smyrnis, Nguyen, Marten, Wortsman, Ghosh, Zhang, et~al.]{datacomp}
Samir~Yitzhak Gadre, Gabriel Ilharco, Alex Fang, Jonathan Hayase, Georgios Smyrnis, Thao Nguyen, Ryan Marten, Mitchell Wortsman, Dhruba Ghosh, Jieyu Zhang, et~al.
\newblock Datacomp: In search of the next generation of multimodal datasets.
\newblock \emph{Advances in Neural Information Processing Systems}, 36, 2024.

\bibitem[Ghosh et~al.(2024)Ghosh, Hajishirzi, and Schmidt]{geneval}
Dhruba Ghosh, Hannaneh Hajishirzi, and Ludwig Schmidt.
\newblock Geneval: An object-focused framework for evaluating text-to-image alignment.
\newblock \emph{Advances in Neural Information Processing Systems}, 36, 2024.

\bibitem[Hudson \& Manning(2019)Hudson and Manning]{gqa}
Drew~A Hudson and Christopher~D Manning.
\newblock Gqa: A new dataset for real-world visual reasoning and compositional question answering.
\newblock In \emph{Proceedings of the IEEE/CVF conference on computer vision and pattern recognition}, pp.\  6700--6709, 2019.

\bibitem[Jayasumana et~al.(2024)Jayasumana, Ramalingam, Veit, Glasner, Chakrabarti, and Kumar]{FID}
Sadeep Jayasumana, Srikumar Ramalingam, Andreas Veit, Daniel Glasner, Ayan Chakrabarti, and Sanjiv Kumar.
\newblock Rethinking fid: Towards a better evaluation metric for image generation.
\newblock In \emph{Proceedings of the IEEE/CVF Conference on Computer Vision and Pattern Recognition}, pp.\  9307--9315, 2024.

\bibitem[Joshi et~al.(2017)Joshi, Choi, Weld, and Zettlemoyer]{joshi2017triviaqa}
Mandar Joshi, Eunsol Choi, Daniel~S Weld, and Luke Zettlemoyer.
\newblock Triviaqa: A large scale distantly supervised challenge dataset for reading comprehension.
\newblock \emph{arXiv preprint arXiv:1705.03551}, 2017.

\bibitem[Lai et~al.(2024)Lai, Zhang, Zhang, Wu, Bai, Timofeev, Du, Gan, Shan, Chuah, Yang, and Cao]{veclip}
Zhengfeng Lai, Haotian Zhang, Bowen Zhang, Wentao Wu, Haoping Bai, Aleksei Timofeev, Xianzhi Du, Zhe Gan, Jiulong Shan, Chen-Nee Chuah, Yinfei Yang, and Meng Cao.
\newblock Veclip: Improving clip training via visual-enriched captions, 2024.
\newblock URL \url{https://arxiv.org/abs/2310.07699}.

\bibitem[Li et~al.(2024{\natexlab{a}})Li, Zhang, Zhang, Guo, Zhang, Zhang, Li, Liu, and Li]{llavanext-ablations}
Bo~Li, Hao Zhang, Kaichen Zhang, Dong Guo, Yuanhan Zhang, Renrui Zhang, Feng Li, Ziwei Liu, and Chunyuan Li.
\newblock Llava-next: What else influences visual instruction tuning beyond data?, May 2024{\natexlab{a}}.
\newblock URL \url{https://llava-vl.github.io/blog/2024-05-25-llava-next-ablations/}.

\bibitem[Li et~al.(2023{\natexlab{a}})Li, Wang, Wang, Ge, Ge, and Shan]{seed}
Bohao Li, Rui Wang, Guangzhi Wang, Yuying Ge, Yixiao Ge, and Ying Shan.
\newblock Seed-bench: Benchmarking multimodal llms with generative comprehension.
\newblock \emph{arXiv preprint arXiv:2307.16125}, 2023{\natexlab{a}}.

\bibitem[Li et~al.(2022{\natexlab{a}})Li, Li, Xiong, and Hoi]{blip}
Junnan Li, Dongxu Li, Caiming Xiong, and Steven Hoi.
\newblock Blip: Bootstrapping language-image pre-training for unified vision-language understanding and generation.
\newblock In \emph{International conference on machine learning}, pp.\  12888--12900. PMLR, 2022{\natexlab{a}}.

\bibitem[Li et~al.(2022{\natexlab{b}})Li, Li, Xiong, and Hoi]{li2022blip}
Junnan Li, Dongxu Li, Caiming Xiong, and Steven Hoi.
\newblock Blip: Bootstrapping language-image pre-training for unified vision-language understanding and generation.
\newblock In \emph{International conference on machine learning}, pp.\  12888--12900. PMLR, 2022{\natexlab{b}}.

\bibitem[Li et~al.(2024{\natexlab{b}})Li, Tu, Hui, Wang, Zhao, Xiao, Ren, Mei, Liu, Zheng, et~al.]{li2024if}
Xianhang Li, Haoqin Tu, Mude Hui, Zeyu Wang, Bingchen Zhao, Junfei Xiao, Sucheng Ren, Jieru Mei, Qing Liu, Huangjie Zheng, et~al.
\newblock What if we recaption billions of web images with llama-3?
\newblock \emph{arXiv preprint arXiv:2406.08478}, 2024{\natexlab{b}}.

\bibitem[Li et~al.(2023{\natexlab{b}})Li, Du, Zhou, Wang, Zhao, and Wen]{pope}
Yifan Li, Yifan Du, Kun Zhou, Jinpeng Wang, Xin Zhao, and Ji-Rong Wen.
\newblock Evaluating object hallucination in large vision-language models.
\newblock In \emph{EMNLP}, 2023{\natexlab{b}}.

\bibitem[Liu et~al.(2023{\natexlab{a}})Liu, Lin, Li, Wang, Yacoob, and Wang]{liu2023mitigating}
Fuxiao Liu, Kevin Lin, Linjie Li, Jianfeng Wang, Yaser Yacoob, and Lijuan Wang.
\newblock Mitigating hallucination in large multi-modal models via robust instruction tuning.
\newblock In \emph{The Twelfth International Conference on Learning Representations}, 2023{\natexlab{a}}.

\bibitem[Liu et~al.(2023{\natexlab{b}})Liu, Li, Wu, and Lee]{llava}
Haotian Liu, Chunyuan Li, Qingyang Wu, and Yong~Jae Lee.
\newblock Visual instruction tuning, 2023{\natexlab{b}}.

\bibitem[Lu et~al.(2022)Lu, Mishra, Xia, Qiu, Chang, Zhu, Tafjord, Clark, and Kalyan]{sciqa}
Pan Lu, Swaroop Mishra, Tanglin Xia, Liang Qiu, Kai-Wei Chang, Song-Chun Zhu, Oyvind Tafjord, Peter Clark, and Ashwin Kalyan.
\newblock Learn to explain: Multimodal reasoning via thought chains for science question answering.
\newblock \emph{Advances in Neural Information Processing Systems}, 35:\penalty0 2507--2521, 2022.

\bibitem[Lu et~al.(2023)Lu, Bansal, Xia, Liu, Li, Hajishirzi, Cheng, Chang, Galley, and Gao]{lu2023mathvista}
Pan Lu, Hritik Bansal, Tony Xia, Jiacheng Liu, Chunyuan Li, Hannaneh Hajishirzi, Hao Cheng, Kai-Wei Chang, Michel Galley, and Jianfeng Gao.
\newblock Mathvista: Evaluating mathematical reasoning of foundation models in visual contexts.
\newblock \emph{arXiv preprint arXiv:2310.02255}, 2023.

\bibitem[McKinzie et~al.(2024)McKinzie, Gan, Fauconnier, Dodge, Zhang, Dufter, Shah, Du, Peng, Weers, et~al.]{mm1}
Brandon McKinzie, Zhe Gan, Jean-Philippe Fauconnier, Sam Dodge, Bowen Zhang, Philipp Dufter, Dhruti Shah, Xianzhi Du, Futang Peng, Floris Weers, et~al.
\newblock Mm1: Methods, analysis \& insights from multimodal llm pre-training.
\newblock \emph{arXiv preprint arXiv:2403.09611}, 2024.

\bibitem[Paperno et~al.(2016)Paperno, Kruszewski, Lazaridou, Pham, Bernardi, Pezzelle, Baroni, Boleda, and Fern{\'a}ndez]{paperno2016lambada}
Denis Paperno, Germ{\'a}n Kruszewski, Angeliki Lazaridou, Quan~Ngoc Pham, Raffaella Bernardi, Sandro Pezzelle, Marco Baroni, Gemma Boleda, and Raquel Fern{\'a}ndez.
\newblock The lambada dataset: Word prediction requiring a broad discourse context.
\newblock \emph{arXiv preprint arXiv:1606.06031}, 2016.

\bibitem[Peebles \& Xie(2023)Peebles and Xie]{dit}
William Peebles and Saining Xie.
\newblock Scalable diffusion models with transformers.
\newblock In \emph{Proceedings of the IEEE/CVF International Conference on Computer Vision}, pp.\  4195--4205, 2023.

\bibitem[Podell et~al.(2023)Podell, English, Lacey, Blattmann, Dockhorn, M{\"u}ller, Penna, and Rombach]{SDXL}
Dustin Podell, Zion English, Kyle Lacey, Andreas Blattmann, Tim Dockhorn, Jonas M{\"u}ller, Joe Penna, and Robin Rombach.
\newblock Sdxl: Improving latent diffusion models for high-resolution image synthesis.
\newblock \emph{arXiv preprint arXiv:2307.01952}, 2023.

\bibitem[Radford et~al.(2021)Radford, Kim, Hallacy, Ramesh, Goh, Agarwal, Sastry, Askell, Mishkin, Clark, et~al.]{CLIP}
Alec Radford, Jong~Wook Kim, Chris Hallacy, Aditya Ramesh, Gabriel Goh, Sandhini Agarwal, Girish Sastry, Amanda Askell, Pamela Mishkin, Jack Clark, et~al.
\newblock Learning transferable visual models from natural language supervision.
\newblock In \emph{ICML}, pp.\  8748--8763, 2021.

\bibitem[Raffel et~al.(2020)Raffel, Shazeer, Roberts, Lee, Narang, Matena, Zhou, Li, and Liu]{raffel2020exploring}
Colin Raffel, Noam Shazeer, Adam Roberts, Katherine Lee, Sharan Narang, Michael Matena, Yanqi Zhou, Wei Li, and Peter~J Liu.
\newblock Exploring the limits of transfer learning with a unified text-to-text transformer.
\newblock \emph{Journal of machine learning research}, 21\penalty0 (140):\penalty0 1--67, 2020.

\bibitem[Rohrbach et~al.(2018)Rohrbach, Hendricks, Burns, Darrell, and Saenko]{CHAIR}
Anna Rohrbach, Lisa~Anne Hendricks, Kaylee Burns, Trevor Darrell, and Kate Saenko.
\newblock Object hallucination in image captioning.
\newblock \emph{EMNLP}, 2018.

\bibitem[Rombach et~al.(2022)Rombach, Blattmann, Lorenz, Esser, and Ommer]{SD}
Robin Rombach, Andreas Blattmann, Dominik Lorenz, Patrick Esser, and Bj{\"o}rn Ommer.
\newblock High-resolution image synthesis with latent diffusion models.
\newblock In \emph{Proceedings of the IEEE/CVF conference on computer vision and pattern recognition}, pp.\  10684--10695, 2022.

\bibitem[Sakaguchi et~al.(2021)Sakaguchi, Bras, Bhagavatula, and Choi]{sakaguchi2021winogrande}
Keisuke Sakaguchi, Ronan~Le Bras, Chandra Bhagavatula, and Yejin Choi.
\newblock Winogrande: An adversarial winograd schema challenge at scale.
\newblock \emph{Communications of the ACM}, 64\penalty0 (9):\penalty0 99--106, 2021.

\bibitem[Schuhmann et~al.(2021)Schuhmann, Vencu, Beaumont, Kaczmarczyk, Mullis, Katta, Coombes, Jitsev, and Komatsuzaki]{laion}
Christoph Schuhmann, Richard Vencu, Romain Beaumont, Robert Kaczmarczyk, Clayton Mullis, Aarush Katta, Theo Coombes, Jenia Jitsev, and Aran Komatsuzaki.
\newblock Laion-400m: Open dataset of clip-filtered 400 million image-text pairs.
\newblock \emph{arXiv preprint arXiv:2111.02114}, 2021.

\bibitem[Singh et~al.(2019{\natexlab{a}})Singh, Natarajan, Shah, Jiang, Chen, Batra, Parikh, and Rohrbach]{singh2019towards}
Amanpreet Singh, Vivek Natarajan, Meet Shah, Yu~Jiang, Xinlei Chen, Dhruv Batra, Devi Parikh, and Marcus Rohrbach.
\newblock Towards vqa models that can read.
\newblock In \emph{CVPR}, 2019{\natexlab{a}}.

\bibitem[Singh et~al.(2019{\natexlab{b}})Singh, Natarajan, Shah, Jiang, Chen, Batra, Parikh, and Rohrbach]{textvqa}
Amanpreet Singh, Vivek Natarajan, Meet Shah, Yu~Jiang, Xinlei Chen, Dhruv Batra, Devi Parikh, and Marcus Rohrbach.
\newblock Towards vqa models that can read.
\newblock In \emph{Proceedings of the IEEE/CVF conference on computer vision and pattern recognition}, pp.\  8317--8326, 2019{\natexlab{b}}.

\bibitem[Sun et~al.(2023)Sun, Shen, Cao, Liu, Li, Shen, Gan, Gui, Wang, Yang, et~al.]{sun2023aligning}
Zhiqing Sun, Sheng Shen, Shengcao Cao, Haotian Liu, Chunyuan Li, Yikang Shen, Chuang Gan, Liang-Yan Gui, Yu-Xiong Wang, Yiming Yang, et~al.
\newblock Aligning large multimodal models with factually augmented rlhf.
\newblock \emph{arXiv preprint arXiv:2309.14525}, 2023.

\bibitem[Tong et~al.(2024)Tong, Brown, Wu, Woo, Middepogu, Akula, Yang, Yang, Iyer, Pan, et~al.]{cambrian}
Shengbang Tong, Ellis Brown, Penghao Wu, Sanghyun Woo, Manoj Middepogu, Sai~Charitha Akula, Jihan Yang, Shusheng Yang, Adithya Iyer, Xichen Pan, et~al.
\newblock Cambrian-1: A fully open, vision-centric exploration of multimodal llms.
\newblock \emph{arXiv preprint arXiv:2406.16860}, 2024.

\bibitem[Wang et~al.(2023)Wang, Lv, Yu, Hong, Qi, Wang, Ji, Yang, Zhao, Song, Xu, Xu, Li, Dong, Ding, and Tang]{cogvlm}
Weihan Wang, Qingsong Lv, Wenmeng Yu, Wenyi Hong, Ji~Qi, Yan Wang, Junhui Ji, Zhuoyi Yang, Lei Zhao, Xixuan Song, Jiazheng Xu, Bin Xu, Juanzi Li, Yuxiao Dong, Ming Ding, and Jie Tang.
\newblock Cogvlm: Visual expert for pretrained language models, 2023.

\bibitem[Welbl et~al.(2017)Welbl, Liu, and Gardner]{welbl2017crowdsourcing}
Johannes Welbl, Nelson~F Liu, and Matt Gardner.
\newblock Crowdsourcing multiple choice science questions.
\newblock \emph{arXiv preprint arXiv:1707.06209}, 2017.

\bibitem[Yu et~al.(2024{\natexlab{a}})Yu, Yao, Zhang, He, Han, Cui, Hu, Liu, Zheng, Sun, et~al.]{yu2024rlhf}
Tianyu Yu, Yuan Yao, Haoye Zhang, Taiwen He, Yifeng Han, Ganqu Cui, Jinyi Hu, Zhiyuan Liu, Hai-Tao Zheng, Maosong Sun, et~al.
\newblock Rlhf-v: Towards trustworthy mllms via behavior alignment from fine-grained correctional human feedback.
\newblock In \emph{Proceedings of the IEEE/CVF Conference on Computer Vision and Pattern Recognition}, pp.\  13807--13816, 2024{\natexlab{a}}.

\bibitem[Yu et~al.(2024{\natexlab{b}})Yu, Yang, Li, Wang, Lin, Liu, Wang, and Wang]{mmvet}
Weihao Yu, Zhengyuan Yang, Linjie Li, Jianfeng Wang, Kevin Lin, Zicheng Liu, Xinchao Wang, and Lijuan Wang.
\newblock Mm-vet: Evaluating large multimodal models for integrated capabilities.
\newblock In \emph{Forty-first International Conference on Machine Learning}, 2024{\natexlab{b}}.

\bibitem[Yue et~al.(2023)Yue, Ni, Zhang, Zheng, Liu, Zhang, Stevens, Jiang, Ren, Sun, et~al.]{yue2023mmmu}
Xiang Yue, Yuansheng Ni, Kai Zhang, Tianyu Zheng, Ruoqi Liu, Ge~Zhang, Samuel Stevens, Dongfu Jiang, Weiming Ren, Yuxuan Sun, et~al.
\newblock Mmmu: A massive multi-discipline multimodal understanding and reasoning benchmark for expert agi.
\newblock \emph{arXiv preprint arXiv:2311.16502}, 2023.

\bibitem[Zellers et~al.(2019)Zellers, Holtzman, Bisk, Farhadi, and Choi]{zellers2019hellaswag}
Rowan Zellers, Ari Holtzman, Yonatan Bisk, Ali Farhadi, and Yejin Choi.
\newblock Hellaswag: Can a machine really finish your sentence?
\newblock \emph{arXiv preprint arXiv:1905.07830}, 2019.

\bibitem[Zhai et~al.(2022)Zhai, Wang, Mustafa, Steiner, Keysers, Kolesnikov, and Beyer]{lit}
Xiaohua Zhai, Xiao Wang, Basil Mustafa, Andreas Steiner, Daniel Keysers, Alexander Kolesnikov, and Lucas Beyer.
\newblock Lit: Zero-shot transfer with locked-image text tuning.
\newblock In \emph{Proceedings of the IEEE/CVF conference on computer vision and pattern recognition}, pp.\  18123--18133, 2022.

\bibitem[Zhang et~al.(2024{\natexlab{a}})Zhang, Zhang, Dong, Zang, and Wang]{zhang2024longclip}
Beichen Zhang, Pan Zhang, Xiaoyi Dong, Yuhang Zang, and Jiaqi Wang.
\newblock Long-clip: Unlocking the long-text capability of clip.
\newblock \emph{ECCV}, 2024{\natexlab{a}}.

\bibitem[Zhang et~al.(2024{\natexlab{b}})Zhang, Gao, Gan, Dufter, Wenzel, Huang, Shah, Du, Zhang, Li, Dodge, You, Yang, Timofeev, Xu, Chen, Fauconnier, Lai, You, Wang, Dehghan, Grasch, and Yang]{zhang2024mm15methodsanalysis}
Haotian Zhang, Mingfei Gao, Zhe Gan, Philipp Dufter, Nina Wenzel, Forrest Huang, Dhruti Shah, Xianzhi Du, Bowen Zhang, Yanghao Li, Sam Dodge, Keen You, Zhen Yang, Aleksei Timofeev, Mingze Xu, Hong-You Chen, Jean-Philippe Fauconnier, Zhengfeng Lai, Haoxuan You, Zirui Wang, Afshin Dehghan, Peter Grasch, and Yinfei Yang.
\newblock Mm1.5: Methods, analysis \& insights from multimodal llm fine-tuning.
\newblock \emph{arXiv preprint arXiv:2409.20566}, 2024{\natexlab{b}}.

\bibitem[Zhu et~al.(2023)Zhu, Chen, Shen, Li, and Elhoseiny]{minigpt}
Deyao Zhu, Jun Chen, Xiaoqian Shen, Xiang Li, and Mohamed Elhoseiny.
\newblock Minigpt-4: Enhancing vision-language understanding with advanced large language models.
\newblock \emph{arXiv preprint arXiv:2304.10592}, 2023.

\end{thebibliography}
\bibliographystyle{iclr2025_conference}

\newpage
\appendix
\appendixpage
\definecolor{darkred}{RGB}{139, 0, 0}
\definecolor{darkblue}{RGB}{0, 0, 139}
\definecolor{darkgreen}{RGB}{0, 139, 0}
\definecolor{darkyellow}{RGB}{153, 153, 0}
\definecolor{darkpurple}{RGB}{102, 0, 102}
\definecolor{darkbrown}{RGB}{102, 51, 0}

\setcounter{figure}{0}
\makeatletter 
\renewcommand{\thefigure}{A\@arabic\c@figure}
\makeatother
\setcounter{table}{0}
\renewcommand{\thetable}{A\arabic{table}}

We provide additional details for datasets, experimental settings, results, and analysis in the supplementary material. 

\section{Experimental Details}

\subsection{CLIP}
We summarize the training details in Table~\ref{tab:hyper}. For the pre-training stage, we pre-train models on up to 512 TPUs with JAX~\citep{jax2018github}. 

\begin{table*}[ht]
    \minipage{1\linewidth}
    \caption{\small Pre-training hyper-parameters and settings for the in-house CLIP.}
    \label{tab:hyper}
    \centering
    \small
    \setlength{\tabcolsep}{2.5pt}
    \renewcommand{\arraystretch}{1.35}
    \begin{tabular}{l|c}
    \toprule
    Batch size & $32768$ \\
    Image size & $224\times 224$ (ViT-B/16) \\
    Image pre-processing & long-side resizing with padding (i.e., \texttt{tf.image.resize\_with\_pad})\\
    Text tokenizer & T5~\citep{raffel2020exploring}, lowercase \\
    Text maximum length & $77$ \\
    Steps & $435,000$ (i.e., $\sim 14$B examples seen) \\
    Optimizer & AdamW ($\beta_1=0.9, \beta_2=0.98$) \\
    Peak learning rate (LR) & $0.0005$\\
    LR schedule & cosine decays with linear warm-up (first $2$k steps)\\
    Weight decay & $0.2$ \\
    Dropout rate & $0.0$ \\
    \bottomrule
    \end{tabular}
    \endminipage
\end{table*}

\subsubsection{Additional Experiments}
To further explore the performance gap between DSC and SSC on CLIP, we present two additional benchmarks to enhance the representativeness of CLIP’s existing evaluations: 1) linear probing and 2) transferability between CLIP pre-trained with different captions and LLaVA-style MLLMs. Linear probing provides a direct measure of the quality and generalization capability of the representations learned by CLIP. Strong performance from a linear classifier on specific tasks indicates that the pre-trained model has effectively captured relevant and discriminative features, underscoring the robustness of its embeddings. Additionally, we assess CLIP’s representation quality using LLaVA~\citep{llava} as a case study, where the vision encoder remains frozen during both pre-training and SFT stages. This makes LLaVA an ideal benchmark for evaluating the transferability and integrity of CLIP’s learned representations. 

We summarize the results on linear probing in Fig.~\ref{tab:linear_probing}: even though DSC and SSC shows lower zero-shot performance, they achieve comparable results to AltText after linear probing, indicating similar pre-trained representations. Furthermore, combining synthetic captions with AltText yields the best overall performance. Then we use these pre-trained vision encoders and insert them into LLaVA and complete the default pre-training and SFT stages in LLaVA~\citep{llava}. All of our pre-trained CLIPs use ViT-B/16 as the backbone. We use Vicuna-1.3 as the LLM for LLaVA training and report recent benchmarks in Table~\ref{tab:llava}: POPE~\citep{pope}, TextVQA~\citep{textvqa}, GQA~\citep{gqa}, SciQA~\citep{sciqa}, LLaVA-Bench~\citep{llava}, MME~\citep{mme}, and MM-Vet~\citep{mmvet}. In this case, the combination of SSC and AltText achieves the highest overall performance, leading in 5 out of 9 columns. This highlights the critical role of both synthetic captions and original AltTexts in CLIP's pre-training: synthetic captions enhance image-text alignment, while AltTexts introduce valuable data diversity.
\begin{table*}[ht]
    \centering
    \caption{\small LLaVA as the benchmark for evaluating CLIP vision encoder pre-trained on different captions.}
    \vspace{-0.2cm}
    \small
    \resizebox{0.95\textwidth}{!}{
    \begin{tabular}{c|c|c|c|cc|c|cc|c}
    \toprule[1.2pt]
\multirow{2}{*}{Pre-trained Caption} & POPE  & \multirow{2}{*}{TextVQA} & \multirow{2}{*}{GQA} & \multicolumn{2}{c|}{SciQA} & LLaVA-Bench & \multicolumn{2}{c|}{MME} & \multirow{2}{*}{MM-Vet} \\
       & Avg. &                   &                   &  IMG       &   Accuracy       & COCO & Perception           &    Cognition      &                   \\
     \midrule
     AltText & 84.4  & \textbf{50.2} & 59.1 & 64.3 & 64.8 & 76.4 & 1332.6 & 271.1 & 25.0 \\
DSC & 83.4 & 47.6 & 59.1 & 63.2 & 64.8 & 75.6 & 1307.8 & \textbf{312.9} & 23.9 \\
DSC + AltText  & 84.6 & 49.8 & 59.9 & 63.6 & 65.3 & 77.2 & \textbf{1405.3} & 273.2 & \textbf{25.5}  \\
SSC & 84.5 & 48.2 & 59.6 & 59.0 & 62.7 & 76.7 & 1343.6 & 248.6 & 22.4 \\
SSC + AltText & \textbf{84.7} & 49.4 & \textbf{60.1} & \textbf{65.3} & \textbf{65.8} & \textbf{77.9} & 1370.7 & 270.7 & 25.1 \\

    \bottomrule[1.2pt]
\end{tabular}
    }
    \label{tab:llava}
\end{table*}
\begin{table}[t]
\begin{tabular}{cc}

\hspace{-0.15cm}%
\begin{minipage}[t]{0.61\linewidth}
\vspace{0pt}
\centering

\caption{\small{Compatibility between rewritten-based and filtering-based methods. We use DFN-2B~\citep{DFN} as the example and train CLIP with ViT/B-16 on different captions. 
}}
\vspace{-0.2cm}
\label{tab:dfn_clip}   
\small
\resizebox{1.0\textwidth}{!}{
  \begin{tabular}{ccccc|cc}
        \toprule[1.2pt]
        \multirow{2}{*}{\bf Pre-train Caption}  & \multicolumn{2}{c}{\bf COCO (R@1)} & \multicolumn{2}{c}{\bf Flickr30k (R@1)} & \multirow{2}{*}{\bf ImageNet} & \multirow{2}{*}{\bf ImageNetV2} \\
         & \bf I-T & \bf T-I & \bf I-T & \bf T-I & & \\
        \midrule
        AltText &61.22	&49.27	&86.20	&68.94	&\textbf{76.12}	&\textbf{68.49}  \\
        DSC  &51.47	&30.99	&77.28	&55.68	&27.97	&23.67 \\
        SSC & 59.06	&31.91	&87.41	&62.49	&53.96	&46.39 \\
        DSC+SSC &60.68	&38.46	&89.60&	68.68&	56.10	&49.15 \\
        AltText+DSC+SSC &\textbf{70.56}	&\textbf{50.74}	&\textbf{92.40}	&\textbf{76.92}	&72.45	&64.98 \\

        \bottomrule[1.2pt]
    \end{tabular}
}

\end{minipage}

&
\hspace{-0.3cm}%
\begin{minipage}[t]{0.38\linewidth}
\vspace{0pt}

\includegraphics[width=\linewidth]{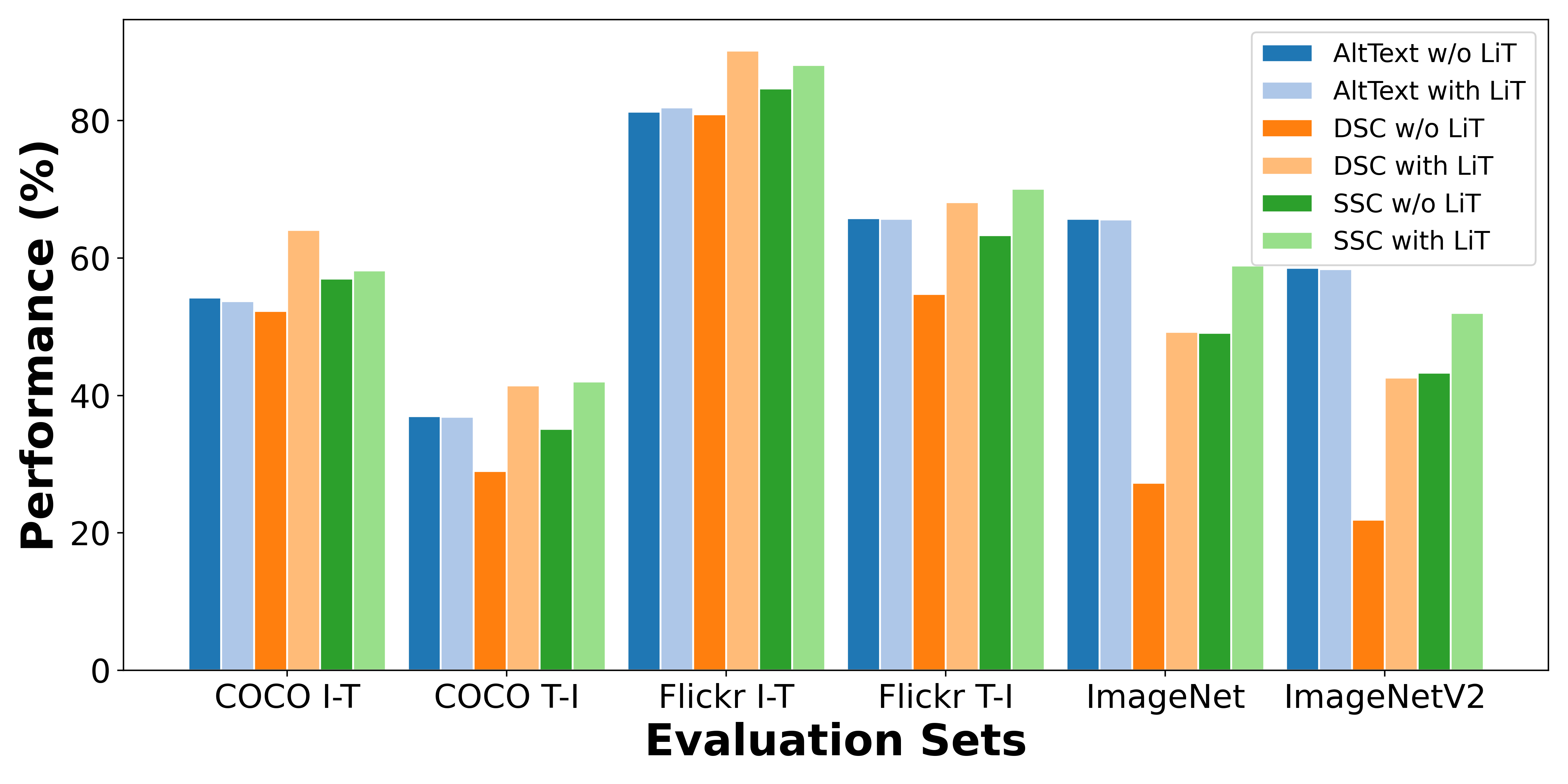}
\vspace{-0.7cm}
\captionof{figure}{\footnotesize Effect of LiT~\citep{lit} on different captions after pre-training.}
\label{fig:lit}

\end{minipage}

\end{tabular}
\vspace{-0.3cm}
\end{table}

\textbf{Compatibility of rewritten-based methods and filtering-based methods.} Besides rewritten-based datasets like web-crawled VeCap-300M~\citep{veclip}, which leverage recaptioning techniques to improve image-text alignment, it is essential to evaluate the compatibility between rewriting-based and filtering-based methods that remove mismatched image-text pairs. We consider DFN-2B~\citep{DFN} as a representative example, where a pre-trained CLIP model filters the dataset to retain pairs that align well with CLIP's capabilities. A key research direction is to explore whether synthetic captions can effectively replace the original CLIP-selected AltText. To investigate this, we apply our captioning pipeline to DFN-2B~\citep{DFN} to generate synthetic captions and pre-train CLIP models on these captions. The results are summarized in Table~\ref{tab:dfn_clip}. In this CLIP-filtered dataset, neither rewritten DSC nor SSC, nor their combination, outperforms the original AltText. However, when combining all our synthetic captions with the original AltText (using uniform sampling during training), we observe significant improvements in retrieval tasks, such as \textbf{+9.34\%} on COCO I-T and \textbf{+7.98\%} on Flickr T-I tasks, demonstrating the enhanced image-text alignment provided by our synthetic captions. Nevertheless, incorporating synthetic captions results in a performance drop of around 4\% on ImageNet, highlighting the crucial role of the diverse information contained in the original AltText for CLIP's learning.

\textbf{Effect of Locked-image text Tuning (LiT)~\citep{lit} after pre-training with different captions.} LiT~\citep{lit} trains a text model to derive meaningful representations from a pre-trained image model by freezing the image encoder and fine-tuning only the text encoder. We investigate the impact of LiT after pre-training CLIP on different captions, with the results summarized in Fig.~\ref{fig:lit}. During the LiT stage, we continue to train the text encoder using AltText. Our findings reveal that LiT with AltText consistently benefits both SSC and TSC across all evaluation sets, highlighting once again the critical role of AltText in CLIP training.

\subsection{Multimodal LLM}
We summarize the training details in Table~\ref{tab:hyper_mllm}. For the pre-training stage, we pre-train models on up to 512 TPUs with JAX~\citep{jax2018github}.

\begin{table*}[h!]
    \minipage{1\linewidth}
    \caption{\small Pre-training hyper-parameters and settings for the Multimodal LLM experiments. We use the same configuration as the 1.2B model in MM1~\citep{mm1}.}
    \label{tab:hyper_mllm}
    \centering
    \small
    \setlength{\tabcolsep}{2.5pt}
    \renewcommand{\arraystretch}{1.35}
    \begin{tabular}{l|c}
    \toprule
    \multicolumn{2}{c}{General}\\
    \midrule
    Batch size & $512$ \\
    Image encoder & $336\times 336$ ViT-L/14\\
    Visual-language connector & C-Abstractor with $144$ image tokens\\
    Language model & $1.2$B transformer decoder-only language model\\
    Steps & $50000$\\
    \bottomrule
    \end{tabular}
    \endminipage
\end{table*}

For the SFT experiments, we follow the same datasets and configuration as in MM1~\citep{mm1}.

\textbf{TextCore.} For the pre-training benchmarks, TextCore is an average number of 8 benchmarks: ARC~\citep{ARC}, PIQA~\citep{bisk2020piqa}, LAMBADA~\citep{paperno2016lambada}, WinoGrande~\citep{sakaguchi2021winogrande},
HellaSWAG~\citep{zellers2019hellaswag}, SciQ~\citep{welbl2017crowdsourcing}, TriviaQA~\citep{joshi2017triviaqa}, and WebQS~\citep{berant2013semantic}.

\textbf{SFT dataset.} 
For the SFT benchmarks, we summarize the details in Table~\ref{tab:mm1_benchmark_details}: we mainly use MME~\citep{mme}, SEED~\citep{seed}, POPE~\citep{pope}, LLaVA-Bench (Wild)~\citep{llava}, MM-Vet~\citep{mmvet}, TextVQA~\citep{singh2019towards}, MMMU~\citep{yue2023mmmu},  MathVista~\citep{lu2023mathvista}, ScienceQA~\citep{sciqa}.

\begin{table}[ht]
    \centering
    \resizebox{0.85\linewidth}{!}{%
    \begin{tabular}{c|c}
         \toprule
         Benchmark & Metric\\
         \midrule
         MME~\citep{mme} & Normalized Accuracy \\
         SEED~\citep{seed} & Seed-IMG\\
         POPE~\citep{pope} & Average of random, popular and adversarial\\
         LLaVA-Bench (Wild)~\citep{llava} & GPT-assisted score\\
         MM-Vet~\citep{mmvet} & GPT-assisted score\\
         TextVQA~\citep{singh2019towards} &VQA Open Flamingo Accuracy\\
         MMMU~\citep{yue2023mmmu} & Accuracy \\
         MathVista~\citep{lu2023mathvista} & GPT-assisted score\\ 
         ScienceQA~\citep{sciqa} &Accuracy-IMG\\
         \bottomrule
    \end{tabular}
    }
    \vspace{1mm}
    \caption{Details of benchmarks and their metrics used in MM1.}
    \label{tab:mm1_benchmark_details}
\end{table}

\subsection{Diffusion Model}
We summarize the training details of our self-implemented diffusion model based on Stable Diffusion 3~\citep{sd3} in Table~\ref{tab:hyper_sd3}. 

\begin{table*}[ht]
    \minipage{1\linewidth}
    \caption{\small Pre-training hyper-parameters for our diffusion model based on Stable Diffusion 3.}
    \label{tab:hyper_sd3}
    \centering
    \small
    \setlength{\tabcolsep}{2.5pt}
    \renewcommand{\arraystretch}{1.35}
    \begin{tabular}{l|c}
    \toprule
    Batch size & $4096$ \\
    Image size & $256\times 256$ \\
    Step & $500,000$ \\
    Text condition & in-house CLIP's G/14 text encoder with T5 tokenizer ~\citep{raffel2020exploring} \\
    Text maximum token length & $77$ \\
    Optimizer & Adafactor ($\beta_1=0.9, \beta_2=0.999$) \\
    Learning rate (LR) & $0.0001$ (constant with linear warm-up for $1$k steps) \\
    Ema decay & $0.9999$ \\
    Classifier free guidance & $7.5$ \\
    \bottomrule
    \end{tabular}
    \endminipage
\end{table*}

\section{A Deeper Analysis of Generated Captions}
Fig.~\ref{fig:captioner} is an overview of our two-stage fine-tuning process: we first convert a MLLM into an image captioner, then we further fine-tune it to convert it into a human-aligned captioner. 
Our smaller image captioning model (3B) efficiently generates large volumes of synthetic data for our experiments. Using this model, we re-captioned a dataset of 7 billion images across multiple iterations. Furthermore, our larger model, with 7 billion parameters, is designed to produce more detailed captions, surpassing the level of detail offered by our long caption format.

\textbf{AFC fine-tuning dataset.} To generate AltText Fusion Captions (AFC), we also prepare a fine-tuning dataset in this format. Specifically, given AltText and a DSC caption generated by our captioner, we ask LLM to fuse AltText information to the DSC. By this way, we construct a 20K training dataset for our captioner.

\textbf{Less hallucinations in our DSC. } 
The Caption Hallucination Assessment with Image Relevance (CHAIR) metric~\citep{CHAIR} is a custom-designed evaluation tool developed to identify and measure the extent of object hallucination in image captioning tasks. The metric determines the proportion of generated words that accurately correspond to objects present in the image, as verified by ground truth sentences and object segmentations. It includes two scores: one that measures the fraction of hallucinated object instances (referred to as $\text{CHAIR}_{i}$), and the other that calculates the fraction of sentences containing at least one hallucinated object (referred to as $\text{CHAIR}_{s}$): 
\begin{equation*}
\begin{aligned}
    \text{CHAIR}_{i} &= \frac{\vert\{\text{hallucinated objects}\}\vert}{\vert\{\text{all objects mentioned}\}\vert}, & 
    \text{CHAIR}_{s} &= \frac{\vert\{\text{sentences with hallucinated object}\}\vert}{\vert\{\text{all sentences}\}\vert}.
\end{aligned}
\end{equation*}


       
    



\begin{wraptable}{r}{0.5\textwidth}
\vspace{-0.3cm}
    \centering
    \caption{\small{Hallucination detection across different MLLMs and our captioner.}}
    \vspace{-0.2cm}
    \label{tab:chair}   
    \small
    \resizebox{0.45\textwidth}{!}{ 
    \begin{tabular}{c|cc}
        \toprule[1.2pt]
        Model & $\text{CHAIR}_{i} (\downarrow)$ & $\text{CHAIR}_{s} (\downarrow)$ \\
        \midrule
        InstructBLIP~\citep{instructblip} & 14.5 & 30.0 \\
        MiniGPT-4~\citep{minigpt} & 8.2 & 24.2 \\
        Shikra~\citep{shikra} & 7.0 & 22.0 \\
        LLaVA-1.5~\citep{llava} & 6.2 & 20.6 \\
        \midrule
        \rowcolor{tbgray} \textbf{Our} & \textbf{5.9} & \textbf{19.6} \\
        \bottomrule[1.2pt]
    \end{tabular}
    }
\end{wraptable}
As shown in Table~\ref{tab:chair}, our model achieves a CHAIR${i}$ score of 5.9 and a CHAIR${s}$ score of 19.6, outperforming leading models such as LLaVA-1.5~\citep{llava}, Shikra~\citep{shikra}, and MiniGPT-4~\citep{minigpt}. The lower CHAIR scores indicate that our captioner produces fewer hallucinated objects per instance and fewer sentences containing hallucinated objects. This improvement shows the effectiveness of our two-stage fine-tuning process, which strategically reduces objects hallucination by leveraging human-aligned data and strict prompt constraints. Consequently, our model offers more reliable and accurate objects recognition capabilities, making it a robust tool for generating high-quality captions across various applications.

\begin{figure}[ht]
    \centering
    \includegraphics[width=0.95\linewidth]{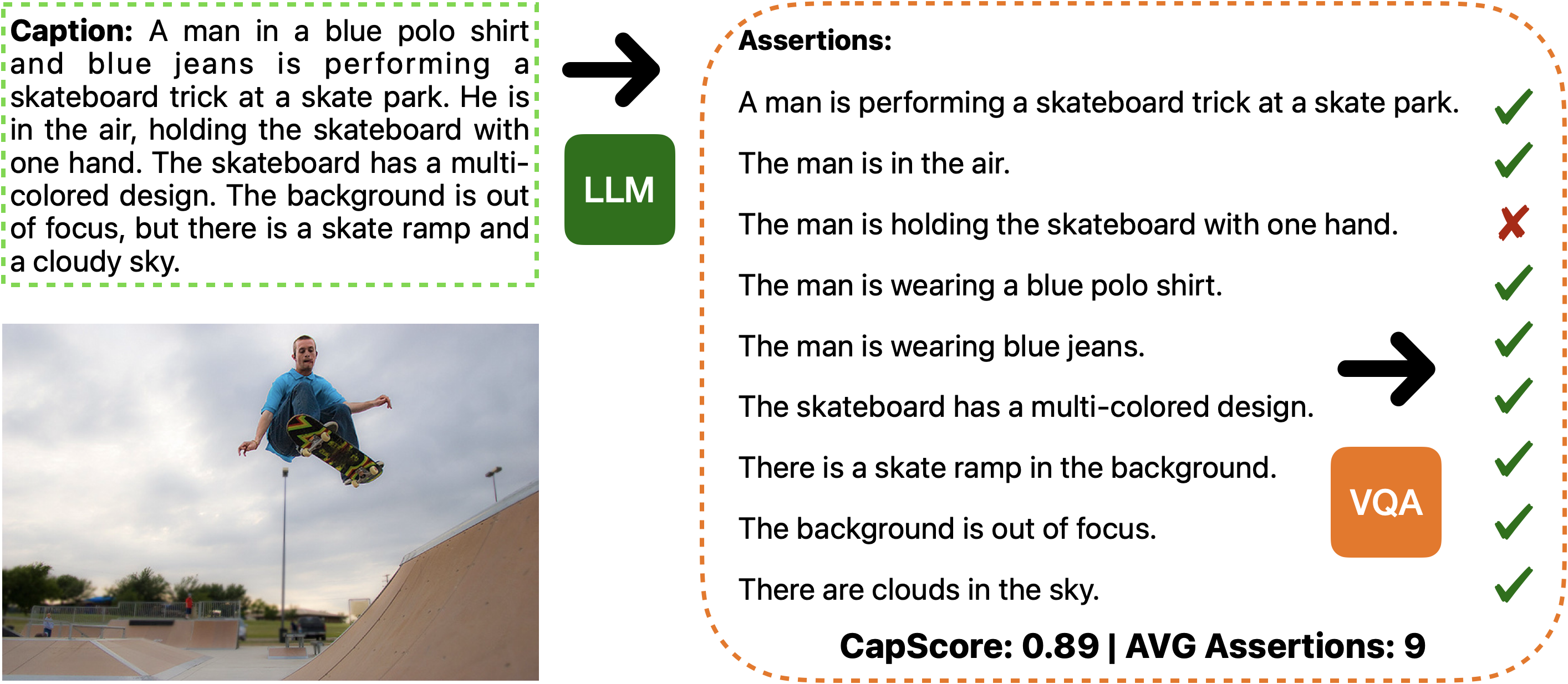}
    \caption{An overview of CapScore to evaluate the quality of captions: we use LLM to generate assertions based on the caption and then use a VQA model to check these assertions. }
    \label{fig:cap_score}
\end{figure}

\textbf{ANA: Average Number of Assertions, a metric for evaluating richness of captions.} Besides hallucination, the richness of captions are also an important index to control the generated captions. We propose ANA to quantify the richness of captions. Inspired by GenEval~\citep{geneval} for evaluating text-to-image generation models, we reverse the process of text-to-image to image-to-text. As shown in Fig.~\ref{fig:cap_score}, we prompt an LLM to generate different assertions of a caption. After that, we can also leverage a VQA model to check if these details are aligned with the visual contents.

\textbf{CapScore: A metric for evaluating hallucinations in synthetic captions.} Although the CHAIR metric is widely used, we find it insufficient for detecting hallucinations in object attributes, especially in highly descriptive captions. To overcome this limitation, we propose a new metric to evaluate both the hallucination and richness of captions. Inspired by GenEval~\citep{geneval} for evaluating text-to-image generation models, we reverse the process of text-to-image to image-to-text and propose CapScore. 
CapScore measures the correctness of synthetic captions by evaluating the alignment between generated textual assertions and the actual content of the image. As shown in  Fig.~\ref{fig:cap_score}, CapScore has two key steps: 1) use an LLM to extract structured assertions from the captions; 2) use a MLLM to serve as a VQA (Visual Question Answering) model to verify the truthfulness of these assertions. Specifically, each assertion represents a distinct factual claim made within the caption. Then the VQA model determines whether the image supports each claim by answering questions based on the assertions. CapScore is then defined as the percentage of assertions validated as correct by the VQA model. A higher CapScore indicates fewer hallucinations and greater factual accuracy in the generated captions. 

As shown in Table~\ref{tab:cap_score}, there is a notable trade-off between the richness of captions and their accuracy. As captions become longer, the Average Number of Assertions (ANA) increases, reflecting the growing richness and complexity of the generated captions. However, CapScore drops with longer captions, which suggests that while the captions provide more content, they are more prone to hallucinations—where the captioning model introduces incorrect or irrelevant information not present in the image. For instance, while DSC+ produces the highest ANA, it also demonstrates the lowest CapScore, highlighting this balance. By contrast, SSC maintains a higher CapScore with fewer assertions, demonstrating better alignment with the image content but at the cost of less detailed descriptions.

This behavior highlights the importance of balancing richness and accuracy in multimodal tasks. Models aiming for high-precision applications (e.g., zero-shot classification with CLIP) may benefit from shorter captions (e.g., SSC), whereas scenarios requiring more detailed scene descriptions (e.g., mutlimodal LLMs) may prioritize longer captions like DSC+, accepting some decrease in factual accuracy.

 

\begin{table}[ht]
\centering
\caption{\small CapScore and Average Number of Assertions (ANA) to evaluate the richness and accuracy of different captions.}
    \vspace{-0.2cm}
    \small
    \resizebox{0.5\textwidth}{!}{
\begin{tabular}{c|c|c}
\toprule[1.2pt]
\textbf{Caption} & $\textbf{CapScore} (\uparrow)$ & $\textbf{ANA} (\uparrow)$ \\
 \midrule 
LLaVA-1.5~\citep{llava} & 88.76 & 7.30\\
SSC &91.56 & 2.49 \\
DSC & 87.30 & 8.13 \\
DSC+ & 75.74 & 12.20  \\
\bottomrule[1.2pt] 
 \end{tabular}
}
\label{tab:cap_score}
    \vspace{-0.1cm} 
\end{table}

\textbf{More detailed but noisy captions are helpful for MLLM pre-training.} We examine the impact of hallucinations in image-text data used for MLLM pre-training, focusing on the type of hallucinations detected in our captions. For our primary comparison, we select the best-performing model, LLaVA-1.5~\citep{llava}, as shown in Table~\ref{tab:chair}. We first utilize LLaVA-1.5 to generate captions on the VeCap-300M dataset~\citep{veclip}. Next, we apply our captioner to generate DSC+ for the same set of images. Our captions are more detailed but contain more hallucinations on object. We use MM1's pre-training under a small scale setting as a case study to examine whether image-caption data with fewer hallucinations or with more details can offer advantages for MLLM pre-training. We evaluate the two models after applying fine-tuning using the same data recipe.

As shown in Table~\ref{tab:captioner_vs_llava}, the MLLM pre-trained with more detailed captions performs better, even though these data contain more hallucinations. When examining task-specific results, we observe that our DSC+ pre-trained model outperforms the LLaVA captions model on 7 out of 9 benchmarks. Specifically, DSC+ improves performance on VQAv2, MMMU, MathV, and SEED, among others, indicating that the added detail from DSC+ benefits these tasks. However, despite the slight degradation in MMEP results (727.2 vs. 773.4), the overall advantage of detailed captions with minor hallucinations demonstrates that a balance between information richness and accuracy can positively impact MLLM's performance. The larger gains in multimodal understanding tasks, such as MMEC and LLaVAW, suggest that hallucination-tolerant MLLM pre-training may help with complex vision-language reasoning.

\begin{table}[ht]
\centering
\caption{\small SFT evaluation of models pre-trained with captions generated by LLaVA-1.5 and our DSC+. We use the same SFT recipe and evaluation benchmarks as in MM1~\citep{mm1}.}
    \vspace{-0.2cm}
    \small
    \resizebox{0.95\textwidth}{!}{
\begin{tabular}{c|ccccccccc}
\toprule[1.2pt]
\textbf{Pre-Trained Data} & \textbf{VQAv2} & \textbf{VQA$^\text{T}$} & \textbf{MMMU} & \textbf{MathV} & \textbf{MME$^\text{P}$} & \textbf{MME$^\text{C}$} & \textbf{SEED} & \textbf{POPE} & \textbf{LLaVA$^\text{W}$} \\
\midrule 
      
LLaVA Captions & 78.0 & 65.4 & 30.0 & 27.6 & \textbf{773.4} & 200.4 & 63.5 & 83.7 & 63.6 \\  
\rowcolor{tbgray} DSC+ & \textbf{79.0} & \textbf{65.4} & \textbf{32.6} & \textbf{29.4} & 727.2 & \textbf{224.3} & \textbf{66.8} & \textbf{85.1} & \textbf{71.5} \\
 \bottomrule[1.2pt] 
 
 \end{tabular}
}
\label{tab:captioner_vs_llava}
    \vspace{-0.1cm} 
\end{table}

\end{document}